\documentclass{article}

\usepackage{microtype}
\usepackage{graphicx}
\usepackage{subfigure}
\usepackage{booktabs} 

\usepackage{hyperref}

\usepackage[accepted]{icml2025}

\usepackage{amsmath}
\usepackage{amssymb}
\usepackage{mathtools}
\usepackage{amsthm}

\usepackage{algorithm}
\usepackage{algorithmic}
\usepackage{amsfonts}
\usepackage{mathrsfs}
\usepackage{multirow}
\usepackage{makecell}
\usepackage{tablefootnote}
\usepackage{threeparttable}
\usepackage{xr}
\usepackage{algorithm}
\usepackage{algorithmic}
\usepackage{diagbox}

\usepackage[capitalize,noabbrev]{cleveref}

\theoremstyle{plain}
\newtheorem{theorem}{Theorem}[section]

\newtheorem{lemma}[theorem]{Lemma}
\newtheorem{corollary}[theorem]{Corollary}
\theoremstyle{definition}
\newtheorem{definition}[theorem]{Definition}
\newtheorem{assumption}[theorem]{Assumption}
\theoremstyle{remark}

\newenvironment{theorem1proof}{\noindent{\textbf{Proof of Theorem \ref{Theorem of the bounds of the mean classification risk}}:}}{\quad \hfill$\Box$}
\newenvironment{corollary1proof}{\noindent{\textbf{Proof of Corollary \ref{Corollary of the upper bound of the classification risk}}:}}{\quad \hfill$\Box$}
\newenvironment{proposition1proof}{\noindent{\textbf{Statement of Assumption \ref{Proposition of the effect of SVD on the labeling error}}:}}{\quad \hfill$\Box$}
\newenvironment{assumption4.1statement}{\noindent{\textbf{Statement of Assumption \ref{Assumption of perfect alignment for f^*}}:}}{\quad \hfill$\Box$}
\newenvironment{corollary2proof}{\noindent{\textbf{Proof of Theorem \ref{Theorem of the classification risk bounds on the optimal function after optimal truncated SVD}}:}}{\quad \hfill$\Box$}
\newenvironment{corollary3proof}{\noindent{\textbf{Proof of Theorem \ref{Theorem of the upper bound of downstream classification error}}:}}{\quad \hfill$\Box$}

\usepackage[textsize=tiny]{todonotes}

\icmltitlerunning{How does Labeling Error Impact Contrastive Learning? A Perspective from Data Dimensionality Reduction}

\begin{document}

\twocolumn[
\icmltitle{How does Labeling Error Impact Contrastive Learning? A Perspective from Data Dimensionality Reduction}



\icmlsetsymbol{equal}{*}

\begin{icmlauthorlist}
\icmlauthor{Jun Chen}{1}
\icmlauthor{Hong Chen}{1,2}
\icmlauthor{Yonghua Yu}{3}
\icmlauthor{Yiming Ying}{4}
\end{icmlauthorlist}

\icmlaffiliation{1}{College of Informatics, Huazhong Agricultural University, Wuhan, China}
\icmlaffiliation{2}{Engineering Research Center of Intelligent Technology for Agriculture, Ministry of Education, China}
\icmlaffiliation{3}{College of Engineering, Huazhong Agricultural University, Wuhan, China}
\icmlaffiliation{4}{School of Mathematics and Statistics, University of Sydney, NSW, Australia}

\icmlcorrespondingauthor{Hong Chen}{chenh@mail.hzau.edu.cn}

\icmlkeywords{Machine Learning, ICML}

\vskip 0.3in
]



\printAffiliationsAndNotice{}  

\begin{abstract}
In recent years, contrastive learning has achieved state-of-the-art performance in the territory of self-supervised representation learning. 
Many previous works have attempted to provide the theoretical understanding underlying the success of contrastive learning. 
Almost all of them rely on a default assumption, i.e., the label consistency assumption, which may not hold in practice (the probability of failure is called labeling error) due to the strength and randomness of common augmentation strategies, such as random resized crop (RRC). 
This paper investigates the theoretical impact of labeling error on the downstream classification performance of contrastive learning. 
We first reveal several significant negative impacts of labeling error on downstream classification risk. 
To mitigate these impacts, data dimensionality reduction method (e.g., singular value decomposition, SVD) is applied on original data to reduce false positive samples, and establish both theoretical and empirical evaluations. 
Moreover, it is also found that SVD acts as a double-edged sword, which may lead to the deterioration of downstream classification accuracy due to the reduced connectivity of the augmentation graph. 
Based on the above observations, we give the augmentation suggestion that we should use some moderate embedding dimension (such as $512, 1024$ in our experiments), data inflation, weak augmentation, and SVD to ensure large graph connectivity and small labeling error to improve model performance.
\end{abstract}

\section{Introduction}
Contrastive learning, as an emerging self-supervised learning paradigm,  has achieved remarkable performance by leveraging data  without label information \cite{DBLP:conf/cvpr/He0WXG20,DBLP:conf/nips/ChenKSNH20,DBLP:conf/cvpr/Jang023,DBLP:conf/nips/WangZCHLYTLWZZ23,DBLP:journals/jmlr/JiDN0Z23}. 
Typically, this learning framework entails formulating  an auxiliary contrastive task endowed with pseudo-labels,
which aims to maximize the similarity between two samples augmented from the same original sample while minimizing the similarity between samples augmented from different original samples \cite{DBLP:conf/icml/ChenK0H20}. 

Recently, some studies have delved into the theoretical mechanism underlying the empirical success of contrastive learning \cite{DBLP:conf/nips/MikolovSCCD13,DBLP:conf/naacl/PagliardiniGJ18,DBLP:conf/aaai/JeanWSALE19,DBLP:conf/icml/AroraKKPS19,DBLP:conf/iclr/JingVLT22,DBLP:conf/iclr/WangZWYL22,DBLP:conf/icml/LeiYYZ23}. 
Generally speaking, they acquiesce that 
the labels of two augmented samples generated from the same original sample remain consistent, which is  referred to as the label consistency assumption \cite{DBLP:conf/iclr/WangZWYL22}. Particularly,  
\citet{DBLP:conf/iclr/WangZWYL22} stated that label consistency is a natural and minimal assumption that is likely to hold in practice, and established the upper and lower bounds of downstream classification risk for contrastive learning  only requiring intra-class samples have similar augmented views (called intra-class augmentation overlap, see Figure \ref{Figure of different labels, intra-class overlap and inter-class overlap} (b)). 
Nevertheless, given that data augmentation process is random, some strong augmentation strategies like RRC may lead to the lose of semantic-related information \cite{DBLP:conf/icml/Zang00Z0L024}, which undermines label consistency and gives rise to labeling error \cite{DBLP:conf/nips/TamkinGHG23}. 
For instance, Figure \ref{Figure of different labels, intra-class overlap and inter-class overlap} (a) shows that several images augmented from a dog image using RRC possess different labels (dog, ship, fog and blanket). 
In such a situation, it can be observed that inter-class samples also exhibit the augmentation overlap phenomenon (see Figure \ref{Figure of different labels, intra-class overlap and inter-class overlap} (c)). 
This inconsistent label phenomenon (called labeling error) motivates us to develop new theoretical analysis for contrastive learning to better understand the interplay among data augmentation, labeling error, and downstream classification performance.

\begin{figure*}[t]
    \centering
    \includegraphics[width=0.8\textwidth]{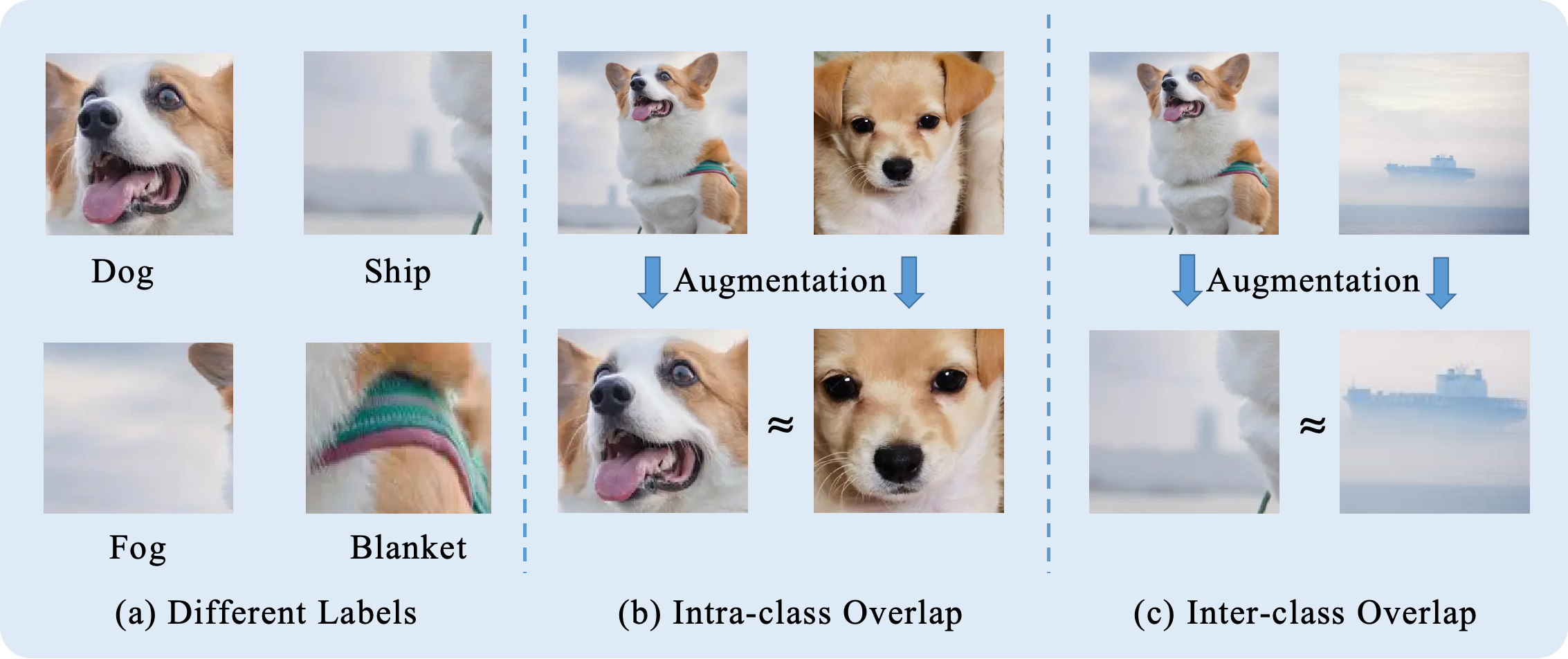}
    \caption{(a) Four images augmented from a single dog image using RRC have different labels (dog, ship, fog and blanket). (b) Augmented
    views of different intra-class samples may overlap. For instance, two views labeled as “dog” might be augmented from different dog images. (c) Augmented views of different inter-class samples may also overlap. For instance, two views labeled as “ship” might be augmented from a dog image and a ship images, respectively.}
    \label{Figure of different labels, intra-class overlap and inter-class overlap}
\end{figure*}

Recently, the analysis framework of \citet{DBLP:conf/nips/HaoChenWGM21} pioneered the consideration of the labeling error caused by data augmentation, which revealed the first dependencies on labeling error and the connectivity of the augmentation graph (Equation (2) in \citet{DBLP:conf/nips/HaoChenWGM21}) for the downstream linear probe error. While matching the experimental observations, 
they didn't offer any suggestions to reduce the dependency on labeling error.
Subsequently, considering that generated data may sometimes even harm contrastive learning, \citet{DBLP:conf/iclr/WangZ024} formulated a novel analysis strategy to explore the  reasons underlying this labeling error phenomenon from the perspective of data inflation
\footnote{Data inflation was defined by \citet{DBLP:conf/iclr/WangZ024} as the process of using generative models (e.g., denoising diffusion probabilistic model (DDPM) \cite{DBLP:conf/nips/HoJA20}) to generate a lot of synthetic samples. \citet{DBLP:conf/iclr/WangZ024} performed contrastive learning on the combination of the real and generated data.}. 
Theoretical and empirical results in \citet{DBLP:conf/iclr/WangZ024} illustrated  that stronger data inflation would bring larger graph connectivity, which decreases the upper bound of the downstream linear probe error. 
They also validated that stronger data augmentation would affect the trade-off of labeling error and graph connectivity to bring the phenomenon that model performance first rises and then falls. 
Consequently, they suggested obtaining large graph connectivity along with small labeling error through the combination of strong data inflation and weak augmentation. 
However, good downstream performance is primarily ensured by large graph connectivity. 
They didn't verify whether the labeling error caused by the weak augmentation is sufficiently small. 

In this paper, we investigate the negative impacts of labeling error. 
Specifically, we first provide the theoretical upper and lower bounds of downstream classification risk and error \footnote{Classification error measures the proportion of misclassified samples. Thus, the summation of classification error and accuracy equal to $100\%$.}
considering both intra-class and inter-class augmentation overlap. 
From the data dimensionality reduction perspective, our theoretical and empirical analyses validate the effectiveness of some useful suggestions for decreasing labeling error to enhance downstream classification performance. 
Our main contributions are summarized as follows. 

\begin{itemize}
    \item \emph{Theoretical guarantees of classification risk for both intra-class and inter-class augmentation overlap cases.}
    Beyond the intra-class augmentation overlap considered by \citet{DBLP:conf/iclr/WangZWYL22}, we establish the first upper and lower bounds of downstream classification risk for inter-class augmentation overlap case. 
    It is discovered from these bounds that there are several significant dependencies associated with labeling error. 
    Following the above analysis, a new perspective of data dimensionality reduction is introduced to reduce these dependencies. 
    As illustrated by singular value decomposition (SVD), both theoretical analysis and empirical observations demonstrate that this dimensionality reduction technique is capable of suppressing classification risk by diminishing labeling error. 

    \item \emph{Theoretical guarantees of classification error while using SVD.}
    Except for downstream classification risk, we provide an upper bound of classification error while using SVD.
    This bound uncovers that SVD can also reduce classification error by decreasing labeling error. 
    Besides, SVD might lead to small graph connectivity, thereby potentially worsening this bound. 
    We suggest that adopting a moderate
    \footnote{The word “moderate” means that the embedding dimension $k$ is not too small or too large. 
    In our experiments, $k=512$ or $k=1024$ may be suitable.}
    embedding dimension contributes to the decrease of classification error by increasing graph connectivity, which is validated by our data experiments. 

    \item \emph{Some useful suggestions guided by our analyses.}
    Under the experiment setting of \citet{DBLP:conf/iclr/WangZ024}, our empirical evaluations indicate that their weak data augmentation still brings an unignorable labeling error. 
    The aforementioned analyses suggest that  we should use moderate embedding dimension, data inflation, weak augmentation, and SVD to achieve two critical objectives: ensuring large graph connectivity and small labeling error, thereby enhancing the downstream classification accuracy.
\end{itemize}

\section{Related Works}
\textbf{Theoretical Understanding of Contrastive Learning.} 
Except for the above theoretical studies, there are also many understanding of contrastive learning from other perspectives \cite{DBLP:conf/nips/Tian0PKSI20,DBLP:journals/jmlr/ChenDL20,DBLP:conf/icml/ZimmermannSSBB21,DBLP:conf/icml/SaunshiAGMZAKK22,DBLP:conf/pkdd/WaidaWANZK23,DBLP:journals/jmlr/JiDN0Z23,DBLP:conf/icml/ZhangW023,DBLP:journals/jmlr/ZouL23,DBLP:conf/ijcai/WenLGC24}. 
From the perspective of information theory, 
\citet{DBLP:conf/nips/Tian0PKSI20} theoretically and empirically showed that data augmentation can decrease mutual information and improve downstream classification accuracy. 
\citet{DBLP:conf/icml/SaunshiAGMZAKK22} empirically presented that different function classes and algorithms bring different behaviors and suggested the consideration of inductive biases in theoretical analysis. 
\citet{DBLP:journals/jmlr/JiDN0Z23} proved two conclusions: 1) contrastive learning outperforms some other self-supervised learning paradigms such as autoencoder and generative adversarial network; 2) label information benefits in-domain downstream task and harms out-domain downstream task. 
Different from previous theories on data augmentation \cite{DBLP:conf/icml/DaoGRSSR19,DBLP:conf/icml/RajputFCLP19,DBLP:conf/icml/WuZVR20}, \citet{DBLP:journals/jmlr/ChenDL20} pioneered a connection between data augmentation and the performance of downstream task using group theory under approximate equality condition (similar to label consistency). 
These works all hinge upon the label consistency assumption as a fundamental premise. 
In contrast, our analysis is free from the label consistency assumption.

\textbf{Applications of Contrastive Learning.}
 Contrastive learning has achieve the empirical success  in many fields, including time series prediction \cite{DBLP:conf/icml/NonnenmacherOSR22,DBLP:conf/iclr/LeePL24,DBLP:conf/iclr/XuMWMR24,DBLP:conf/iclr/ZhengW0MC0L24}, graph learning \cite{DBLP:conf/icml/XiaWWCL22,DBLP:conf/aistats/Ghose0HC23,DBLP:conf/nips/LinXDZW23,DBLP:conf/nips/YuWZLS23,DBLP:conf/icml/LiuT024}, federated learning \cite{DBLP:conf/iclr/YuLWXL23,DBLP:conf/iclr/LouizosRK24}, multi-modal learning \cite{DBLP:conf/coling/LinZWSW022,DBLP:conf/nips/WangZC23,DBLP:conf/icip/HuangWJZZ23,DBLP:journals/tkde/FangZHWX23,DBLP:journals/tip/XiaWGYG23}, adversarial learning \cite{DBLP:conf/nips/XuZLSK23,DBLP:conf/nips/ZhangSC0C23,DBLP:conf/nips/XuZLSK23a,DBLP:conf/iclr/Luo0023}, etc. 
Not only self-supervised contrastive learning, there are also some supervised works \cite{DBLP:conf/nips/KhoslaTWSTIMLK20,DBLP:conf/iclr/BarbanoDTGG23} and weak supervised progresses \cite{DBLP:conf/iccv/Zheng0Y0Z0021,DBLP:conf/iclr/TsaiLLLSM22,DBLP:conf/icml/Cui0W023} that combine available label information to guide model training.

\section{Preliminaries}
\label{Preliminaries}
This section initiates our analysis with a comprehensive overview of contrastive learning. 
We first denote $\bar{D}$ the finite but exponentially large set set of all unlabeled original sample $\bar{x} \in \mathbb{R}^d$, and denote $y: \bar{D}\rightarrow [K]$ the ground-truth (deterministic) labeling function, then $y_{\bar{x}}\in \{1,...,K\}$. 
Let $\mathcal{P}$ be the population distribution of the original samples.
Contrastive learning initializes and trains the model parameters in an unsupervised fashion, laying a foundational understanding of the underlying structure and patterns of original data. 
The subsequent supervised fine-tuning stage further adapts the pre-trained model parameters to some specific downstream task on the test data drawing from $\mathcal{P}$ to enhance model performance. 

\textbf{For the first learning stage}, an encoder projector $f \in \mathcal{F}_1: \mathbb{R}^d \rightarrow \mathbb{S}^{k-1}$ is pre-trained to map a $d$-dimensional input vector $\bar{x}$ to an embedding vector $z=f(x)$ in a $k$-dimensional unit hypersphere, where the inequality $k < d$ generally holds which is validated by our experiments. 
This pre-training process is composed with several steps including data augmentation, contrastive representation, and loss calculation. 
Due to the unavailability of true label, contrastive pre-training constructs a surrogate task via some sample augmentation strategies. 
Then, this task is learned by minimizing the distance between similar samples and maximizing the distance between dissimilar samples in embedding space. 
For example, we first select an original sample $\bar{x}$, and make data augmentations $t, t^+\in \mathcal{T} (\mathcal{T}=\{t|t: \mathbb{R}^d \rightarrow \mathbb{R}^d\})$ to obtain two independent augmented samples $x=t(\bar{x}), x^+=t^+(\bar{x})$, respectively. 
We use $p(\cdot|\bar{x})$ to denote the distribution of the augmented sample for $\bar{x}$. 
Then, $M$ negative samples $x_i^-, i=1,...,M$, are randomly augmented from other original samples. 
We assume $\bar{x}, x, x^+, x^-$ follow the same population distribution, which is the same as previous work like \citet{DBLP:conf/iclr/WangZWYL22}. 
We use these augmented samples to calculate the most common contrastive loss, InfoNCE loss \cite{DBLP:journals/corr/OordLV18}, defined as follows
\begin{align}
\label{InfoNCE loss}
    & \mathcal{L}_{InfoNCE}(f)\notag\\
    = & \mathbb{E}_{x,x^+,\{x_i^-\}_{i=1}^M} \left[-\log \frac{e^{f(x)^\top f(x^+)}}{e^{f(x)^\top f(x^+)} + \sum_{i=1}^M e^{f(x)^\top f(x_i^-)}}\right]. 
\end{align}

In InfoNCE loss, the similarity between two samples is quantified by the inner product $f(x)^\top f(x^\prime)$ of two vectors ($x^\prime$ denotes any sample in $x^+, x_1^-, ..., x_M^-$). 
Considering that $f$ is a projector mapping from $\mathbb{R}^d$ to a $k$-dimensional unit hypersphere, the inner product $f(x)^\top f(x^\prime)$ represents cosine similarity. 
To sum up, minimizing InfoNCE loss is equivalent to maximizing the similarity of positive pair and minimizing the similarity of negative pairs.
Generally, we consider the empirical version of InfoNCE loss
\begin{align}
\label{Empirical InfoNCE loss}
    & \hat{\mathcal{L}}_{InfoNCE}(f)\notag \\
    = & - \frac{1}{n_1} \sum\limits_{j=1}^{n_1} \log \frac{e^{f(x_j)^\top f(x_j^+)}}{e^{f(x_j)^\top f(x_j^+)} + \sum_{i=1}^M e^{f(x_j)^\top f(x_{ji}^-)}}
\end{align}
and its minimizer 
\begin{equation*}
    f^*\in \arg\min_{f\in \mathcal{F}_1} \hat{\mathcal{L}}_{InfoNCE}(f).
\end{equation*}

\textbf{For the second learning stage}, the parameters of the pre-trained encoder remain unaltered. We retrain a linear projection head $g\in \mathcal{F}_2: \mathbb{R}^{k} \rightarrow \mathbb{R}^K$ with the weight $W\in \mathbb{R}^{k\times K}$ using the cross entropy (CE) loss to conduct downstream classification task. 
For a labeled sample $x\sim \mathcal{P}$, the CE loss and the mean CE loss 
\footnote{Mean classifier was first considered by \citet{DBLP:conf/icml/AroraKKPS19}.
They stated that the mean classifier could achieve comparable performance to learned weights.
Note that we don't use the mean classifier in our experiments.
It is only available as an intermediate result (Theorem \ref{Theorem of the bounds of the mean classification risk}) in our theoretical analysis.}
are calculated via 
\begin{align}
\label{CE loss}
    \mathcal{L}_{CE}(g_{f,W}) = \mathbb{E}_{x} \left[-\log \frac{\exp\left(f(x)^\top w_{y_{x}}\right)}{\sum_{i=1}^K \exp\left(f(x)^\top w_i\right)}\right]
\end{align}
and 
\begin{align}
\label{Mean CE loss}
    \mathcal{L}_{CE}(g_{f,\mu}) = \mathbb{E}_{x} \left[-\log \frac{\exp\left(f(x)^\top \mu_{y_{x}}\right)}{\sum_{i=1}^K \exp\left(f(x)^\top \mu_i\right)}\right],
\end{align}
where the linear classifier predicts $g_{f,W} = \arg\max_{i\in[K]} \left(f^\top W\right)_i, W=[w_1, ..., w_K]$, and the parameter of mean projection head is $\mu=[\mu_1, ..., \mu_K]$ whose element $\mu_i$ denotes the mean of the representations for the inputs with the label $i\in[K]$, i.e., $\mu_i=\mathbb{E}_{\{x|y_x = i\}}[f(x)]$. Define 
the downstream classification error as
\begin{equation}\label{downstream classification error}
    \mathcal{E}(f, W) = \underset{x\sim \mathcal{P}}{\mathrm{Pr}}\left[g_{f,W}(x) \neq y_{x}\right].
\end{equation}
This work uses linear probing in fine-tuning stage, rather than full fine-tuning which updates all model parameters.

As previously stated, contrastive learning employs data augmentation methods to make preparation for unsupervised pre-training. 
There are two default assumptions: 1) the positive samples pair $(x, x^+)$ have the same label, i.e., $y_x = y_{x^+}$; 2) any negative sample $x_i^-$ has a label different from $y_x$. 
However, the inherent randomness of both traditional data augmentations and negative sample sampling may violate these two assumptions in practice. 
As a consequence, the pre-training process may be led astray, potentially undermining both the overall training effectiveness and the quality of the learned representations. 
This study primarily focuses on the potential labeling error caused by false positive augmented samples (Assumption \ref{Assumption of labeling error}). 

\begin{assumption}[Labeling Error \cite{DBLP:conf/iclr/WangZ024}]
\label{Assumption of labeling error}
    For any $\bar{x}\sim \mathcal{P}$, its latent label $y_{\bar{x}}$, and its augmented sample $x\sim p(\cdot|\bar{x})$, we assume that the true label of $x$ is not consistent with $y_{\bar{x}}$ with the probability $\alpha\in (0, 1)$. 
    That is, $\mathbb{E}_{\bar{x}\sim \mathcal{P}, x\sim p(\cdot|\bar{x})} \left[\mathbb{I}\left[y_x\neq y_{\bar{x}}\right]\right] = \alpha$.
\end{assumption}

\section{Main Results}

\subsection{Theoretical Impact of Labeling Error}
\begin{definition}[Augmentation Overlap]
\label{Definition of augmentation overlap}
    Given a collection of augmentation strategies $\mathcal{T}$, we say that two original samples $\bar{x}, \bar{x}^\prime \sim \mathcal{P}$ are $\mathcal{T}$-augmentation overlapped if they have overlapped views, i.e., $\exists t, t^\prime\in \mathcal{T}$ such that $t(\bar{x})=t^\prime(\bar{x}^\prime)$.
\end{definition}

In the analysis of \citet{DBLP:conf/iclr/WangZWYL22}, they proposed the concept of augmentation overlap. 
Owing to the label consistency, the positive augmented sample pair $(x,x^+)$ has the same label $y_x = y_{x^+}$. 
Therefore, they analyzed the model performance in the case that different intra-class samples could have overlapped augmented views (Figure \ref{Figure of different labels, intra-class overlap and inter-class overlap} (b)). 
While under Assumption \ref{Assumption of labeling error}, there arises the phenomenon that different inter-class samples may also exhibit overlapped augmented views (Figure \ref{Figure of different labels, intra-class overlap and inter-class overlap} (c)). 
Our first theorem (Theorem \ref{Theorem of the bounds of the mean classification risk}) takes these two augmentation overlap cases into account (the proof can be found in \emph{Appendix \ref{theorem1proof}}).


\begin{theorem}[Bounds of Mean Classification Risk]
\label{Theorem of the bounds of the mean classification risk}
    Let Assumption \ref{Assumption of labeling error} hold. 
    For any $f\in \mathcal{F}_1, g\in \mathcal{F}_2$, the gap between the mean downstream classification risk and the contrastive risk $\mathcal{L}_{CE}(g_{f,\mu}) - \mathcal{L}_{InfoNCE}(f)$ can be upper bounded by
    \begin{align*}
        \sqrt{V\left(f(x)\right)} + \sqrt{V^-\left(f(x)\right)} + \mathcal{O}\left(M^{-\frac{1}{2}}\right) - \log\left(\frac{M}{K}\right)
    \end{align*}
    and lower bounded by 
    \begin{align*}
        & - \sqrt{V\left(f(x)\right)} - \sqrt{V^-\left(f(x)\right)} - \frac{1}{2} V(f(x^-))\\
        & - \mathcal{O}\left(M^{-\frac{1}{2}}\right) - \log\left(\frac{M+1}{K}\right),
    \end{align*}
    where $V\left(f(x)\right) = \mathbb{E}_{(x,x^+)\in X^+}\left[\left\|f(x)-\mu_{y_{x}}\right\|^2\right]$, $V^-\left(f(x)\right) = \mathbb{E}_{(x,x^+)\in X^-} \left[\left\|f(x^+)-\mu_{y_{x}}\right\|^2\right]$, $V(f(x^-)) = V(z|z\in \{f(x^+), f(x^-)\}, y_{x^+}=y_x) = \mathbb{E}_{\{z|z\in \{f(x^+), f(x^-)\}, y_{x^+}=y_x\}} \left[\left\|z - \mu_{y_{z}} \right\|^2\right]$ \footnote{$\mu_{y_x}$ and $\mu_{y_z}$ denote the mean of these representations with the same label as sample $x$ and embedding $z$, respectively.} are the intra-class variance of the representations for true positive augmented samples, the variance for false positive augmented samples, and the intra-class variance for negative and true positive augmented samples, respectively, $X^+ = \left\{(x,x^+)| y_x = y_{x^+}\right\}$ and $X^- = \left\{(x,x^+)| y_x \neq y_{x^+}\right\}$ denote the sets of true positive sample pair and false positive sample pair, respectively.
\end{theorem}

\begin{figure*}[ht]
    \centering
    \includegraphics[width=0.8\textwidth]{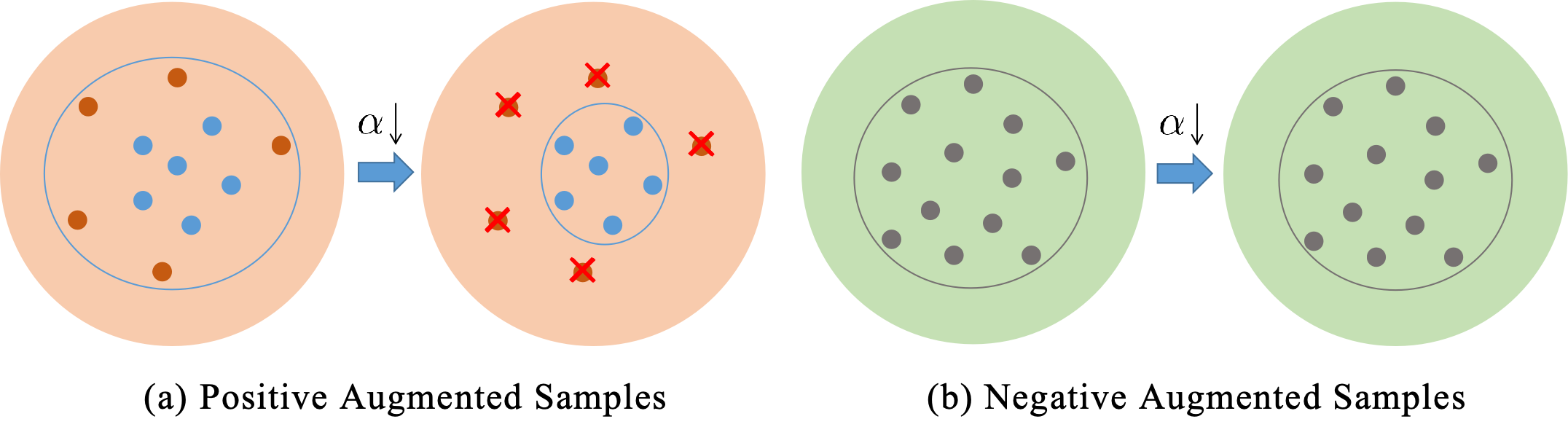}
    \caption{The impact of the labeling error $\alpha$ on the positive (a) and negative (b) augmented samples. Brown dots, blue dots and grey dots denote the false positive samples, true positive samples and true negative samples, respectively. Blue arrows denote the decrease of $\alpha$.}
    \label{Figure of comparisons of the positive and negative augmented samples after SVD}
\end{figure*}

From the above results, it can be observed that there are multiple terms determining the upper and lower bounds of the mean classification risk $\mathcal{L}_{CE}(g_{f,\mu})$, i.e., 
\emph{1)} $\mathcal{L}_{InfoNCE}(f)$: the contrastive risk; 
\emph{2)} $V\left(f(x)\right)$: the intra-class variance of the representations of true positive augmented samples; 
\emph{3)} $V^-\left(f(x)\right)$: the variance of false positive augmented samples; 
\emph{4)} $V(f(x^-)) = V(z|z\in \{f(x^+), f(x^-)\}, y_{x^+}=y_x)$: the intra-class variance of negative and true positive augmented samples;
\emph{5)} $\mathcal{O}\left(M^{-\frac{1}{2}}\right)$: the order of the approximation error; 
\emph{6)} $\log\left(\frac{M}{K}\right)$: a constant depending on $M$ and $K$. 
In terms of the form of these bounds, our result is more refined than that of Theorem 4.2 in \citet{DBLP:conf/iclr/WangZWYL22} as evidenced by these terms $V\left(f(x)\right), V^-\left(f(x)\right)$, and $V(f(x^-))$. 
It should be noted that the assumption of our Theorem \ref{Theorem of the bounds of the mean classification risk} is milder than the corresponding result in \cite{DBLP:conf/iclr/WangZWYL22}, which leads to some latent differences in these terms. 
We carry out the following analyses with respect to these terms. 

\textbf{(1) $V\left(f(x)\right)$:}
Although this term looks like the variance term in the bound of Theorem 4.2 for \citet{DBLP:conf/iclr/WangZWYL22}, its definition $\mathbb{E}_{(x,x^+)\in X^+}\left[\left\|f(x)-\mu_{y_{x}}\right\|^2\right]$ demonstrates the clustering property of true positive sample pair, not including false positive sample pair. 
If the value of $\alpha$ eqals to zero, the variance term in \citet{DBLP:conf/iclr/WangZWYL22} will be equivalent to our $V\left(f(x)\right)$, as shown in Figure \ref{Figure of comparisons of the positive and negative augmented samples after SVD} (a).

\textbf{(2) $V^-\left(f(x)\right)$:}
\citet{DBLP:conf/iclr/WangZWYL22} mistook false positive augmented samples for true positive augmented samples, consequently resulting in a large conditional positive intra-class variance. 
Upon revisiting the definition $\mu_{y_{x}}=\mathbb{E}_{\{x|y=y_x\}}[f(x)]$, these false positive augmented samples do not have an impact on $\mu_{y_{x}}$.
When we discard these false positive augmented samples, the remaining positive augmented samples appear more concentrated around $\mu_{y_{\bar{x}}}$ in comparison to those of \citet{DBLP:conf/iclr/WangZWYL22}. 
Therefore, as the value of $\alpha$ decreases, this term will progressively decrease until it reaches 0.


\textbf{(3) $V(f(x^-))$:}
Considering the sampling randomness and the unavailability of the true labels for negative samples, we let the label $y^-$ denote the latent label of the negative augmented sample $x^-$, irrespective of the class to which its original sample $\bar{x}^-$ pertains.
Therefore, compared with $V^-\left(f(x)\right)$, this term remains unaffected by the labeling error $\alpha$.   
We provide Figure \ref{Figure of comparisons of the positive and negative augmented samples after SVD} to facilitate the understanding of the difference between the analyses of $V^-\left(f(x)\right)$ and $V(f(x^-))$.


To sum up, Theorem \ref{Theorem of the bounds of the mean classification risk} provides more detailed and practical bounds of the mean downstream classification risk $\mathcal{L}_{CE}(g_{f,\mu})$ under milder assumption compared to the results in \citet{DBLP:conf/iclr/WangZWYL22}. 
Our analysis indicates that the labeling error $\alpha$ exerts negative impacts on the theoretical downstream classification risk.
Thus, we aim to put forward some suggestions to mitigate these impacts. 

\subsection{Dimensionality Reduction as A New Perspective}
This subsection investigates the impact of labeling error on contrastive learning through a new perspective of data dimensionality reduction.
Specifically, we use data dimensionality reduction on individual original sample to obtain the corresponding compressed sample which is then augmented to pre-train the model. 
We take Singular Value Decomposition (SVD, \citet{Wall2003}) as an illustrative example. 
For ease of calculation, the randomized SVD (RSVD, \citet{DBLP:journals/siamrev/HalkoMT11}) algorithm is employed in our experiments.

\begin{definition}[SVD]
\label{Definition of SVD}
    For a matrix $X\in\mathbb{R}^{m\times m^\prime}$ $($without loss of generality, we let $m\leq m^\prime)$, its SVD is expressed as $X=USV^\top$, where $U=[u_1,...,u_m]\in \mathbb{R}^{m\times m} (V=[v_1,...,v_{m^\prime}]\in \mathbb{R}^{m^\prime\times m^\prime})$ is the left $($right$)$ singular matrix, consisting of $m (m^\prime)$ orthonormal column vectors $($eigenvectors of $XX^\top (X^\top X))$, $S=[diag(s_1,...,s_m), \mathbf{0}]$ is composed of a diagonal matrix $diag(s_1,...,s_m)\in \mathbb{R}^{m\times m}$ and a zero matrix $\mathbf{0}$ with size $m\times (m^\prime-m)$, the values $s_i (i=1,...,m)$ are the singular value, arranged in descending order such that $s_1\geq s_2\geq...\geq s_m\geq 0$.
\end{definition}
In general, we use the truncated version of SVD which is expressed as $\hat{X}_q=U_q S_q V_q^\top$, where $U_q=[u_1,...,u_q]\in\mathbb{R}^{m\times q}, S_q=diag(s_1,...,s_q), V_q=[v_1,...,v_q]\in\mathbb{R}^{m^\prime\times q}, q\in[m]$. 
The data distribution after conducting truncated SVD is written as $\mathcal{P}_q$, then the corresponding labeling error is $\alpha_q = \mathbb{E}_{\bar{x}\sim \mathcal{P}_q, x\sim p(\cdot|\bar{x})}\left[\mathbb{I}[y_x\neq y_{\bar{x}}]\right]$. 
The following lemma provided by \citet{EckartY36} proved that $\hat{X}_q$ is the best least squares lower rank approximation with rank $q$ for a given matrix $X$. 

\begin{table*}[!ht]
\caption{Downstream classification top-1 accuracies $(\%)$ of SimCLR ($\mathcal{L}_{InfoNCE}$) on CIFAR-10 using the truncated SVD which discards two singular values $(s_{i,i+1}$ denotes we discard the $i$-th and the $i+1$-th singular values $s_i, s_{i+1}$ via SVD, $\mathcal{T}_1=\{$Random resize crop (RRC), Color jitter, Random horizontal flip, Random grayscale, Gaussian blur$\}$, bold number indicates the optimal performance of each experimental group$)$.}
\label{Table of any two singular values discarded}
    \centering
    \renewcommand\arraystretch{1.5}
    \begin{tabular}{c|c|cccccccc}
        \hline
        $\mathcal{T}$ & Encoder & $s_{1,2}$ & $s_{2,3}$ & $s_{3, 4}$ & $s_{4, 5}$ & $s_{5, 6}$ & $s_{6, 7}$ & $s_{7, 8}$ & $s_{8, 9}$\\
        \hline
        $\mathcal{T}_1$ & Resnet-18 & 57.31 & 64.14 & 65.63 & 67.20 & 66.93 & 67.77 & 67.93 & 68.44\\
        \hline
        \hline
        $\mathcal{T}$ & Encoder & $s_{9, 10}$ & $s_{10, 11}$ & $s_{11, 12}$ & $s_{12, 13}$ & $s_{15, 16}$ & $s_{21, 22}$ & $s_{31, 32}$ &\\
        \hline
        $\mathcal{T}_1$ & Resnet-18 & 67.97 & 68.08 & 68.59 & 68.96 & 68.61 & 69.20 & \textbf{69.48} &\\
        \hline
    \end{tabular}
\end{table*}

\begin{figure}[!ht]
    \centering
    \includegraphics[width=0.4\textwidth]{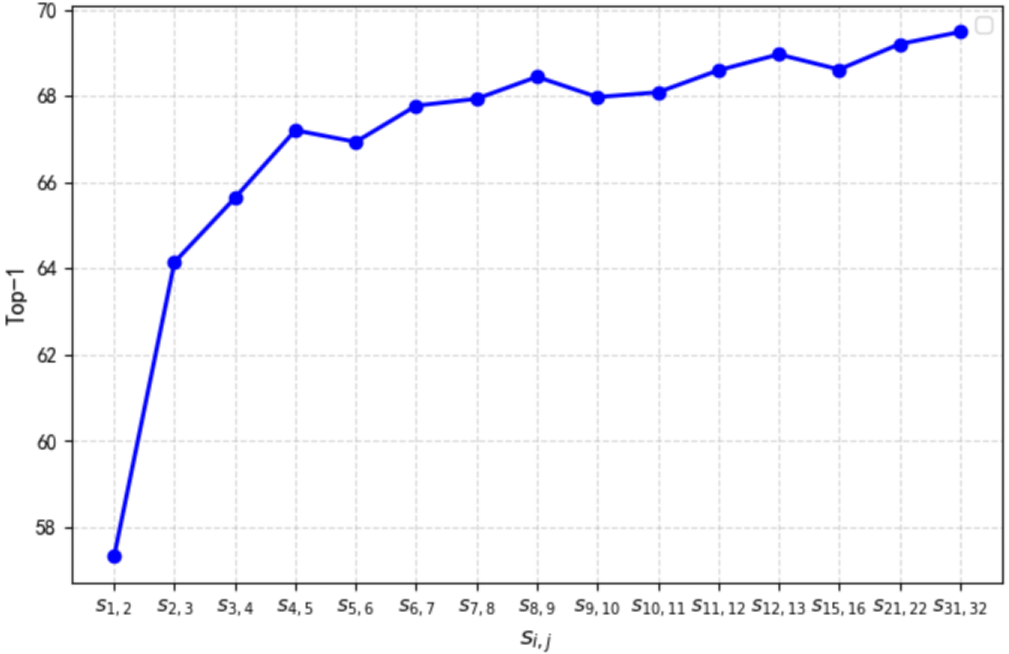}
    \caption{Curve of classification top-1 accuracies ($\%$).}
    \label{Figure of any two singular values discarded}
\end{figure}

\begin{figure}[!ht]
    \centering
    \includegraphics[width=0.4\textwidth]{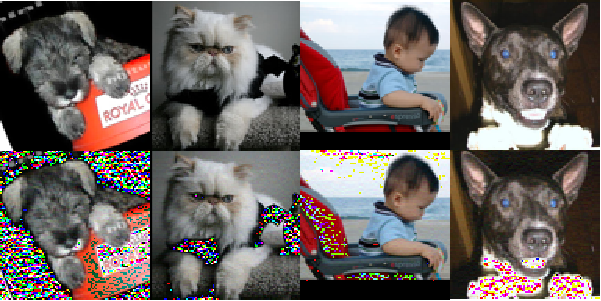}
    \caption{Examples from STL-10. The first row shows the original images and the second row shows the images after taking SVD.}
    \label{Figure of the difference between raw samples and the samples using SVD}
\end{figure}

\begin{lemma}[Eckart-Young Theorem]
\label{Lemma of Eckart-Young Theorem}
    Let $X$ be a $m\times m^\prime$ matrix of rank $r\in[m]$ with complex elements.
    Define $P_q$ as the set of all $m\times m^\prime$ matrices with rank $q\in [r]$. 
    Then, 
    \begin{equation*}
        \left\|X-\hat{X}_q\right\|_F\leq \left\|X-B\right\|_F, \forall B\in P_q.
    \end{equation*}
\end{lemma}
As stated by \citet{KilmerHAN21}, Lemma \ref{Lemma of Eckart-Young Theorem} implies that the majority of the informational content is captured by the dominant singular subspaces, i.e., the span of the singular vectors corresponding to the largest singular values. 
By default, we assume a positive correlation between the amount of information and its semantic relevance. 
It follows that the information content corresponding to the largest singular value $s_1$ should represent the most crucial information in $X$.
For example, an image is intrinsically represented as a matrix $X$. 
The information associated with $s_1$ is most significant for distinguishing the true semantic of the image. 
Table \ref{Table of any two singular values discarded} and Figure \ref{Figure of any two singular values discarded} empirically validate that the larger the singular value is, the more semantic-related information it encompasses. 
Figure \ref{Figure of the difference between raw samples and the samples using SVD} visually presents the differences between original samples and the compressed samples.


\begin{assumption}
\label{Proposition of the effect of SVD on the labeling error}
    Let a sample and its corresponding sample after applying SVD be represented as the matrices $X$ and $\hat{X}_q\in \mathbb{R}^{m\times m^\prime}$, respectively. 
    Assume that there are $q^*$ singular values associated with semantic-related information.
    When $q\geq q^*$, under Assumption \ref{Assumption of labeling error} and the augmentation collection $\mathcal{T}$, the true label of the augmented sample of $\hat{X}_q$ is inconsistent with the latent label of $X$ with the probability $\alpha_q \leq \alpha$.
    Conversely, when $q<q^*$, the corresponding probability satisfies $\alpha_q > \alpha_{q^*}$.
\end{assumption}

\begin{table*}[!ht]
    \centering
    \caption{Downstream classification top-1 accuracies ($\%$) of SimCLR ($\mathcal{L}_{InfoNCE}$) using the truncated SVD with different truncated parameter $q$.}
    \label{Table of w/o SVD vs. SVD under different q}
    \renewcommand\arraystretch{1.5}
    \begin{tabular}{c|c|c|cccccc}
        \hline
        $\mathcal{T}$ & Encoder & Dataset & w/o SVD & $q=30$ & $q=25$ & $q=20$ & $q=15$ & $q=10$\\
        \hline
        $\mathcal{T}_1$ & Resnet-18 & CIFAR-10 & 68.82 & 69.48 & 69.75 & \textbf{69.87} & 69.01 & 68.26\\
        $\mathcal{T}_1$ & Resnet-50 & CIFAR-10 & 63.20 & 63.36 & \textbf{63.96} & 62.23 & 60.97 & 60.06\\
        RRC & Resnet-18 & CIFAR-10 & 58.56 & \textbf{58.83} & 58.67 & 58.61 & 58.54 & 58.32\\
        $\mathcal{T}_1$ & Resnet-18 & CIFAR-100 & 38.48 & 38.81 & \textbf{40.10} & 39.05 & 38.98 & 38.10\\
        \hline
        \hline
        $\mathcal{T}$ & Encoder & Dataset & w/o SVD & $q=90$ & $q=70$ & $q=50$ & $q=30$ & $q=10$\\
        \hline
        $\mathcal{T}_1$ & Resnet-18 & STL-10 & 71.54 & \textbf{73.12} & 72.29 & 71.10 & 70.04 & 67.52\\
        \hline
    \end{tabular}
\end{table*}

\begin{table*}[!ht]
\caption{Downstream classification top-1 accuracies $(\%)$ of SimCLR ($\mathcal{L}_{InfoNCE}$) on CIFAR-10 using the truncated SVD with different augmentations $(\mathcal{T}_2 = \{\mathcal{T}_1 + $ Cutout$\}$; $\mathcal{T}_3=\{$RRC, Cutout, Hide patch$\}$; $\mathcal{T}_4=\{$RRC, Cutout, Color jitter$\}$; $\mathcal{T}_5=\{$RRC, Cutout$\}$; $\mathcal{T}_6=\{$RRC(0.08, 0.5), Cutout$\}$; $\mathcal{T}_7=\{$RRC(0.08, 0.5), Cutout(0.5, 1.0)$\})$.}
\label{Table of w/o SVD vs. SVD under different augmentations}
    \centering
    \setlength{\tabcolsep}{1mm}
    \renewcommand\arraystretch{1.5}
    \begin{tabular}{c|c|ccccccc}
        \hline
        SVD & Encoder & \makebox[0.08\textwidth][c]{$\mathcal{T}_2$} & \makebox[0.08\textwidth][c]{$\mathcal{T}_3$} & \makebox[0.08\textwidth][c]{$\mathcal{T}_4$} & \makebox[0.08\textwidth][c]{$\mathcal{T}_5$} & \makebox[0.08\textwidth][c]{$\mathcal{T}_6$} & \makebox[0.08\textwidth][c]{$\mathcal{T}_7$} & RRC(0.08,0.5)\\
        \hline
        w.o. SVD & Resnet-18 & 62.90 & 50.53 & 60.00 & 56.67 & 54.97 & 54.09 & 57.11\\
        $q=30$ & Resnet-18 & \textbf{64.86} & \textbf{51.00} & \textbf{61.57} & \textbf{57.85} & \textbf{55.69} & \textbf{54.75} & \textbf{58.10}\\
        \hline
    \end{tabular}
\end{table*}

The truncated SVD with $q=q^*$ is defined as the optimal truncated SVD. 
As indicated by Assumption \ref{Proposition of the effect of SVD on the labeling error}, we assume the trend where $\alpha$ initially decreases and then increases as the decrease of $q$. 
In the first stage ($q\geq q^*$), the decrease of $\alpha$ derives from the removal of semantically irrelevant information. 
Once $q<q^*$, semantic-related information begins to be discarded, giving rise to the increase of $\alpha$. 
Table \ref{Table of w/o SVD vs. SVD under different q} and Figure \ref{Classification risk of CIFAR10Resnet18 + CIFAR100Resnet18} (\emph{Appendix \ref{Other Experimental Results}}) present empirical performances applying the truncated SVD with different values of $q$, aligning with our analysis. 
Notably, the values of $q^*$ vary across different settings. 
Table \ref{Table of w/o SVD vs. SVD under different augmentations} further validates the effectiveness of SVD across various augmentation strategies \footnote{Explanations of these augmentation strategies in Table \ref{Table of w/o SVD vs. SVD under different augmentations} are provided in \emph{Appendix \ref{Experimental setting}}}. 

Due to the unavailability of $q^*$, we select a relatively large value of $q$, specifically $q=30$ for CIFAR-10 (image size is $32\times 32$) in Table \ref{Table of w/o SVD vs. SVD under different augmentations}. 
Even though we merely discard the information corresponding to the two smallest singular values $s_{31}, s_{32}$, the results of multiple augmentation strategies in Table \ref{Table of w/o SVD vs. SVD under different augmentations} all exhibit some non-negligible improvements, with an average increase of $0.97\%$.

Further, we introduce an assumption related to SVD, i.e., $(\epsilon_{q^*}, \epsilon_q)$-alignment for any false positive sample pair.

\begin{assumption}[$(\epsilon_{q^*}, \epsilon_q)$-Alignment for Any False Positive Sample Pair]
\label{Assumption of perfect alignment for f^*}
    Let Assumptions \ref{Assumption of labeling error}, \ref{Proposition of the effect of SVD on the labeling error} hold and $X^- = \left\{(x,x^+)| y_x \neq y_{x^+}\right\}$. 
    When performing SVD with the truncated value $q$, the encoder $f$ with the empirical InfoNCE loss $\hat{\mathcal{L}}_{InfoNCE}(f)$ \eqref{Empirical InfoNCE loss} can align any false positive sample pair $(x,x^+)\in X^-, $ such that their distance in the embedding space lies within $[\epsilon(\alpha_{q^*}), \epsilon(\alpha_q)]$, i.e., $\epsilon(\alpha_{q^*}) = \min_{(x,x^+)\in X^-} \|f(x) - f(x^+)\|$ and $\epsilon(\alpha_{q}) = \max_{(x,x^+)\in X^-} \|f(x) - f(x^+)\|$. 
    For simplicity, let $\epsilon_{q^*} = \epsilon(\alpha_{q^*}), \epsilon_q = \epsilon(\alpha_q)$. 
\end{assumption}

Intuitively, labeling error can affect the relationship between the embedding vectors of a positive augmented sample pair. 
Thus, we introduce the $(\epsilon_{q^*}, \epsilon_q)$-alignment assumption, which is inspired by the weak alignment definition proposed by \citet{DBLP:conf/iclr/WangZWYL22} (Definition B.1 in \emph{Appendix B.1}). 
Furthermore, Assumption \ref{Assumption of perfect alignment for f^*} means that the encoder $f$, optimized under the loss $\hat{\mathcal{L}}_{InfoNCE}(f)$, can align any $(x,x^+)\in \left\{(x,x^+)| y_x = y_{x^+}\right\}$ with the distance at most $\epsilon_{q^*}$ in embedding space, that is, $\max_{(x,x^+)\in X^+} \|f(x) - f(x^+)\|\leq \epsilon_{q^*}$. 

Theorem \ref{Theorem of the classification risk bounds on the optimal function after optimal truncated SVD}, stated as below, is derived based on Assumption \ref{Assumption of perfect alignment for f^*}, which theoretically clarifies the effectiveness of data dimensionality reduction.

\begin{theorem}
\label{Theorem of the classification risk bounds on the optimal function after optimal truncated SVD}
        Given the conditions of Theorem \ref{Theorem of the bounds of the mean classification risk} and Assumption \ref{Assumption of perfect alignment for f^*}, after taking the truncated SVD on $\bar{\mathcal{D}}$, the mean downstream classification risk $\mathcal{L}_{CE}(g_{f,\mu}) - \mathcal{L}_{InfoNCE}(f)$ with the encoder $f$ can be upper bounded by 
    \begin{align*}
         \epsilon_{q^*} + \epsilon_q + \mathcal{O}\left(M^{-\frac{1}{2}}\right) - \log\left(\frac{M}{K}\right)
    \end{align*}
    and lower bounded by 
    \begin{align*}
        - \epsilon_{q^*} - \epsilon_{q} - \frac{1}{2} V(f(x^-)) - \mathcal{O}\left(M^{-\frac{1}{2}}\right) - \log\left(\frac{M+1}{K}\right).
    \end{align*}
    When $q=q^*$, the two bounds are $\epsilon_{q^*} + \mathcal{O}\left(M^{-\frac{1}{2}}\right) - \log\left(\frac{M}{K}\right)$ and $- \epsilon_{q^*} - V(f(x^-)) - \mathcal{O}\left(M^{-\frac{1}{2}}\right) - \log\left(\frac{M+1}{K}\right)$.
\end{theorem}

\begin{table*}[ht]
\caption{Downstream classification top-1 accuracies ($\%$) of SimCLR ($\mathcal{L}_{spe}$) on CIFAR-10 using the truncated SVD with different $q$ or the data inflation strategy under the weak data augmentation adopted by \citet{DBLP:conf/iclr/WangZ024} $(\mathcal{T}_8=\{$RRC(0.2, 1.0), Color jitter(0.5, 0.4), Random horizontal flip, Random grayscale, Gaussian blur$\})$.}
\label{Table of weak data augmentation with spectral loss}
    \centering
    \renewcommand\arraystretch{1.5}
    \begin{tabular}{c|c|ccccccc}
        \hline
        $\mathcal{T}$ & Encoder & Inflation & w/o SVD & $q=30$ & $q=25$ & $q=20$ & $q=15$ & $q=10$\\
        \hline
        $\mathcal{T}_8$ & Resnet-18 & 71.54 & 71.21 & 71.64 & \textbf{71.65} & 71.11 & 70.41 & 67.83\\
        \hline
        \hline
        $\mathcal{T}$ & Encoder & Inflation & \multicolumn{2}{c}{Inflation + $(q=30)$} & \multicolumn{2}{c}{Inflation + $(q=25)$} & \multicolumn{2}{c}{Inflation + $(q=20)$}\\
        \hline
        $\mathcal{T}_8$ & Resnet-18 & 71.54 & \multicolumn{2}{c}{71.64} & \multicolumn{2}{c}{\textbf{72.55}} & \multicolumn{2}{c}{71.19}\\
        \hline
    \end{tabular}
\end{table*}

\begin{table*}[ht]
\caption{Downstream classification top-1 accuracies ($\%$) of SimCLR ($\mathcal{L}_{InfoNCE}$) on CIFAR-10 using the truncated SVD with different $q$ or the data inflation strategy under the weak data augmentation adopted by \citet{DBLP:conf/iclr/WangZ024} $(\mathcal{T}_8=\{$RRC(0.2, 1.0), Color jitter(0.5, 0.4), Random horizontal flip, Random grayscale, Gaussian blur$\})$.}
\label{Table of weak data augmentation}
    \centering
    \renewcommand\arraystretch{1.5}
    \begin{tabular}{c|c|ccccccc}
        \hline
        $\mathcal{T}$ & Encoder & Inflation & w/o SVD & $q=30$ & $q=25$ & $q=20$ & $q=15$ & $q=10$\\
        \hline
        $\mathcal{T}_8$ & Resnet-18 & 70.87 & 70.11 & 70.92 & 70.67 & \textbf{70.97} & 70.41 & 69.03\\
        \hline
        \hline
        $\mathcal{T}$ & Encoder & Inflation & \multicolumn{2}{c}{Inflation + $(q=30)$} & \multicolumn{2}{c}{Inflation + $(q=25)$} & \multicolumn{2}{c}{Inflation + $(q=20)$}\\
        \hline
        $\mathcal{T}_8$ & Resnet-18 & 70.87 & \multicolumn{2}{c}{70.98} & \multicolumn{2}{c}{\textbf{71.21}} & \multicolumn{2}{c}{70.48}\\
        \hline
    \end{tabular}
\end{table*}

\subsection{Further Understanding of Labeling Error}
Next, we will delve deeper into the theoretical impact of SVD on the downstream classification error $\mathcal{E}(f,W)$ in \eqref{downstream classification error} through the lens of spectral graph theory. 
In the subsequent analysis, the spectral contrastive loss $\mathcal{L}_{spe}(f)$ is adopted, 
\begin{align*}
    & \mathcal{L}_{spe}(f)\notag\\
    = & -2 \mathbb{E}_{x, x^+} \left[f(x)^\top f(x^+)\right] + \mathbb{E}_{x,x^-}\left[\left(f(x)^\top f(x^-)\right)^2\right],
\end{align*}
which is proposed in \citet{DBLP:conf/nips/HaoChenWGM21} and similar to the InfoNCE loss. 
The augmentation graph, defined as below, is involved in our analysis. 
\begin{definition}[Augmentation Graph, \citet{DBLP:conf/nips/HaoChenWGM21}]
\label{Definition of augmentation graph}
    Given an augmentation collection $\mathcal{T}$, there exist  $n$ augmented samples that form the augmentation dataset \begin{equation*}
        \mathcal{D}_{aug} = \{x|x=t(\bar{x}), \bar{x}\sim \mathcal{P}, t\in \mathcal{T}\}.
    \end{equation*}
    An augmentation graph $\mathcal{G}$ is obtained by taking the $n$ augmented samples as the graph vertices and assuming there exists an edge between two vertices $x, x^\prime \in \mathcal{D}_{aug}$ (if they can be generated from a random original sample $\bar{x}\sim \mathcal{P}$). 

\end{definition}

According to spectral graph theory, we define $A\in \mathbb{R}^{n\times n}$ as the adjacency matrix of the augmentation graph $\mathcal{G}$. 
For two augmented samples $x, x^\prime\in \mathcal{D}_{aug}$, the element $A(x, x^\prime)$ denotes the marginal probability of generating $x, x^\prime$ from a random original sample $\bar{x}\sim \mathcal{P}$. 
Formally, $A(x,x^\prime) = \mathbb{E}_{\bar{x}\sim \mathcal{P}}\left[p(x|\bar{x}) p(x^\prime|\bar{x})\right]$. 
The corresponding normalized graph Laplacian matrix is $L=I-D^{-\frac{1}{2}}AD^{-\frac{1}{2}}$, where $D$ denotes a diagonal degree matrix with the diagonal element $D_{x,x}=\sum_{x^\prime\in\mathcal{D}_{aug}}A(x,x^\prime)$. 
The eigenvalues of $L$ are denoted as $\{\lambda_i\}_{i=1}^n$, where $0=\lambda_1\leq ...\leq \lambda_n\leq 2$. \footnote{When taking the truncated SVD with hyperparameter $q$ on the unlabeled original sample $\bar{x}$, the corresponding symbols are $\mathcal{G}_q, A_q, L_q$, and $\lambda_{i,q}, i \in [n]$.}

\citet{DBLP:conf/nips/HaoChenWGM21} first established an upper bound related to $\alpha$ and $\lambda_{k+1}$ for the downstream classification error of the majority voting classifier
$    \bar{g}_{f,W}(\bar{x}) = \arg\max_{i\in [K]} \underset{x\sim p(\cdot|\bar{x})}{\mathrm{Pr}}\left[g_{f,W}(x) = i\right].$
Building upon their analysis framework, \citet{DBLP:conf/iclr/WangZ024} suggested using strong data inflation and weak data augmentation to guarantee a small value of $\alpha$ and a large value of $\lambda_{k+1}$. 
We aim to replace the complex data inflation with SVD.

Our experimental results in Table \ref{Table of weak data augmentation with spectral loss} and Table \ref{Table of weak data augmentation} demonstrate that the benefit of employing SVD can catch up with that of data inflation. Moreover, we 
theoretically provides the upper bound of the downstream classification error under the application of the truncated SVD with the hyper-parameter $q$.

\begin{table*}[!ht]
\caption{Downstream classification top-1 accuracies ($\%$) of SimCLR ($\mathcal{L}_{spe}$) using the truncated SVD ($q=30$ for CIFAR-10 and CIFAR-100, $q=90$ for STL-10) with different embedding dimension $k$.}
\label{Table of different k with spectral loss}
    \centering
    \renewcommand\arraystretch{1.5}
    \begin{tabular}{c|c|c|ccccc}
        \hline
        \multirow{2}{*}{$\mathcal{T}$} & \multirow{2}{*}{Encoder} & \multirow{2}{*}{Dataset} & \multicolumn{5}{c}{Embedding Dimension}\\
        \cline{4-8}
        & & & $k=128$ & $k=256$ & $k=512$ & $k=1024$ & $k=2048$\\
        \hline
        $\mathcal{T}_1$ & Resnet-18 & CIFAR-10 & 67.71 & 68.51 & 68.54 & \textbf{69.09} & 68.65\\
        $\mathcal{T}_1$ & Resnet-50 & CIFAR-10 & \textbf{67.43} & 65.99 & 66.50 & 66.83 & 66.22\\
        $\mathcal{T}_1$ & Resnet-18 & CIFAR-100 & 35.00 & 36.68 & 36.78 & \textbf{37.78} & 37.18\\
        $\mathcal{T}_1$ & Resnet-50 & CIFAR-100 & 35.46 & 35.42 & 35.39 & \textbf{35.59} & 35.53\\
        $\mathcal{T}_1$ & Resnet-18 & STL-10 & 72.35 & 72.42 & 73.12 & \textbf{73.88} & 73.47\\
        $\mathcal{T}_1$ & Resnet-50 & STL-10 & 74.68 & 74.94 & 75.01 & \textbf{76.26} & 75.57\\
        \hline
    \end{tabular}
\end{table*}

\begin{theorem}[Bounds of Classification Error]
\label{Theorem of the upper bound of downstream classification error}
    Let Assumption \ref{Assumption of labeling error} hold. 
    For the empirical optimal encoder $f^*$ taking the truncated SVD with hyper-parameter $q$ on unlabeled original dataset, there exists a linear head $W^*\in \mathbb{R}^{k\times K}$ with norm $\|W^*\|_F\leq 1/(1-\lambda_{k,q})$ such that 
    \begin{eqnarray*}
        \mathcal{E}(f^*,W^*) \leq \frac{4\alpha_q}{\lambda_{k+1,q}} + 8\alpha_q,
    \end{eqnarray*}
    where $k$ denotes the dimension of embedding space and $\lambda_{k+1, q}$ denotes the $k+1$-th eigenvalues of $L_q$.
\end{theorem}

Note that there are two differences between our Theorem \ref{Theorem of the upper bound of downstream classification error} and Theorem C.3 of \citet{DBLP:conf/nips/HaoChenWGM21}: 
1) we take the truncated SVD on original samples; 
2) we evaluate the classification error for the linear classifier $g_{f,W}$ rather than the majority voting classifier $\bar{g}$
of \citet{DBLP:conf/nips/HaoChenWGM21}.

\citet{DBLP:conf/iclr/WangZ024} concluded that stronger data inflation solely enhances graph connectivity (larger $\lambda_{k+1}$) without affecting labeling error. 
When the graph connectivity is sufficient, only weak data augmentation is necessary to achieve a small labeling error. 
However, Table \ref{Table of weak data augmentation with spectral loss} and Table \ref{Table of weak data augmentation} reveal that the weak augmentation employed by \citet{DBLP:conf/iclr/WangZ024} still induces an non-negligible labeling error. 
Compared with Theorem 4.1 in \citet{DBLP:conf/iclr/WangZ024}, Theorem \ref{Theorem of the upper bound of downstream classification error} further introduces SVD to offer an improvement by reducing $\alpha$ to $\alpha_q$. 
In the following, we will analyze the impact of SVD on  $\alpha_q$ and $\lambda_{k+1, q}$ involved in Theorem \ref{Theorem of the upper bound of downstream classification error}. 

\textbf{(1) $\alpha_q$:}
As discussed behind Assumption \ref{Proposition of the effect of SVD on the labeling error}, $\alpha_q$ exhibits a trend of first decreasing and then increasing as the decrease of $q\in[m]$.
The value of $\alpha_q$ reaches its optimal value $\alpha_{q^*}$ when $q=q^*$.


\textbf{(2) $\lambda_{k+1, q}$:}
As mentioned earlier, $\left\{\lambda_i \right\}_{i=1}^n$ are the eigenvalues of the normalized graph Laplacian matrix $L$. 
Based on the definition of $L$, it can be known that $\left\{1 -\lambda_{i} \right\}_{i=1}^n$ are the eigenvalues of the adjacency matrix $A$, where $1-\lambda_{k+1}$ is the $k+1$-th largest eigenvalue. 
The definition of $A(x,x^\prime)$ demonstrates that the implementation of SVD on original samples will not lead to a decrease in the elements on the main diagonal of $A$. 
In other words, the trace $\mathrm{tr}(A)$ is non-decreasing and remains less than $1$, formally, $\mathrm{tr}(A)=\sum_{i=1}^n \left(1-\lambda_{i,q}\right) \leq 1$. 
Therefore, the application of SVD may lead to $\lambda_{k+1,q}\leq \lambda_{k+1}$, deteriorating the bound of Theorem \ref{Theorem of the upper bound of downstream classification error}. 
It appears that choosing a large dimension $k$ of embedding space may alleviate this deterioration as much as possible by increasing graph connectivity. 
When the value of $k$ is not sufficiently large, such as $k\leq 512$, the corresponding results in Table \ref{Table of different k with spectral loss} may reveal this trend. 
However, Table \ref{Table of different k with spectral loss} also indicates that an overly large value of $k$ might cause the degeneration of the downstream classification performance. 
Interestingly, similar optimal values of $k$ have been observed across experiments on different datasets (CIFAR-10, CIFAR-100 and STL-10) and different encoders (Resnet-18 and Resnet-50). 
Experimental results with InfoNCE loss (Table \ref{Table of different k} in \emph{Appendix \ref{Other Experimental Results}}) also reveal this phenomenon. 
Therefore, there are still some results that do not align with our analysis. 
As stated in Limitation (\emph{Appendix \ref{Limitations}}), we leave the exploration of the underlying reasons for future work. 
In this paper, we propose selecting a moderate value of $k$, such as 512 or 1024. 
Additionally, we recommend adopting data inflation similar to that in \citet{DBLP:conf/iclr/WangZ024} to mitigate the negative impact of SVD on $\lambda$, which is validated by the experiments in the second rows of Table \ref{Table of weak data augmentation with spectral loss} and Table \ref{Table of weak data augmentation}.

In summary, an augmentation strategy is proposed and validated  to ensure large graph connectivity and small labeling error, where the key building blocks include using  moderate embedding dimension, data inflation, weak augmentation and SVD.


\section{Conclusions}
This paper investigated theoretically the impact of labeling error on the downstream classification performance of contrastive learning. 
The derived upper and lower bounds of the downstream classification risk  are both affected by the labeling error. 
To mitigate these negative impacts, we first propose removing the semantically irrelevant information of the original data from the perspective of data dimensionality reduction. 
Specifically, the classical SVD method is employed to offer both theoretical and empirical evidence to support the effectiveness of dimensionality reduction. 
Except for the advantages of conducting SVD on the original data, we also theoretically find that SVD may cause the deterioration of downstream classification accuracy by decreasing the graph connectivity of augmentation graph. 
Based on the aforementioned analysis, we provide an augmentation strategy that we should use moderate embedding dimension (such as $k=512, 1024$), data inflation, weak augmentation and SVD to ensure large graph connectivity and small labeling error, ultimately improving model performance.

\section*{Impact Statement}
This paper presents work whose goal is to advance the field of 
Machine Learning. There are many potential societal consequences 
of our work, none which we feel must be specifically highlighted here.

\section*{Acknowledgments}
This work is supported by the National Natural Science Foundation of China (NSFC) (Nos. 62376104 and 12426512) and the Open Research Fund of Engineering Research Center of Intelligent Technology for Agriculture, Ministry of Education (No. ERCITA-KF002).


\bibliography{References}

\begin{thebibliography}{61}
\providecommand{\natexlab}[1]{#1}
\providecommand{\url}[1]{\texttt{#1}}
\expandafter\ifx\csname urlstyle\endcsname\relax
  \providecommand{\doi}[1]{doi: #1}\else
  \providecommand{\doi}{doi: \begingroup \urlstyle{rm}\Url}\fi

\bibitem[Arora et~al.(2019)Arora, Khandeparkar, Khodak, Plevrakis, and Saunshi]{DBLP:conf/icml/AroraKKPS19}
Arora, S., Khandeparkar, H., Khodak, M., Plevrakis, O., and Saunshi, N.
\newblock A theoretical analysis of contrastive unsupervised representation learning.
\newblock In \emph{International Conference on Machine Learning (ICML)}, volume~97, pp.\  5628--5637, 2019.

\bibitem[Barbano et~al.(2023)Barbano, Dufumier, Tartaglione, Grangetto, and Gori]{DBLP:conf/iclr/BarbanoDTGG23}
Barbano, C.~A., Dufumier, B., Tartaglione, E., Grangetto, M., and Gori, P.
\newblock Unbiased supervised contrastive learning.
\newblock In \emph{International Conference on Learning Representations (ICLR)}, 2023.

\bibitem[Budimir et~al.(2000)Budimir, Dragomir, and Pecaric]{Budimir2000FurtherRR}
Budimir, I., Dragomir, S.~S., and Pecaric, J.
\newblock Further reverse results for jensen's discrete inequality and applications in information theory.
\newblock \emph{Journal of Inequalities in Pure \& Applied Mathematics}, 2, 2000.

\bibitem[Chen et~al.(2020{\natexlab{a}})Chen, Dobriban, and Lee]{DBLP:journals/jmlr/ChenDL20}
Chen, S., Dobriban, E., and Lee, J.~H.
\newblock A group-theoretic framework for data augmentation.
\newblock \emph{Journal of Machine Learning Research}, 21\penalty0 (245):\penalty0 1--71, 2020{\natexlab{a}}.

\bibitem[Chen et~al.(2020{\natexlab{b}})Chen, Kornblith, Norouzi, and Hinton]{DBLP:conf/icml/ChenK0H20}
Chen, T., Kornblith, S., Norouzi, M., and Hinton, G.~E.
\newblock A simple framework for contrastive learning of visual representations.
\newblock In \emph{Proceedings of the 37th International Conference on Machine Learning (ICML)}, volume 119, pp.\  1597--1607, 2020{\natexlab{b}}.

\bibitem[Chen et~al.(2020{\natexlab{c}})Chen, Kornblith, Swersky, Norouzi, and Hinton]{DBLP:conf/nips/ChenKSNH20}
Chen, T., Kornblith, S., Swersky, K., Norouzi, M., and Hinton, G.~E.
\newblock Big self-supervised models are strong semi-supervised learners.
\newblock In \emph{Advances in Neural Information Processing Systems (NeurIPS)}, 2020{\natexlab{c}}.

\bibitem[Cui et~al.(2023)Cui, Huang, Wang, and Wang]{DBLP:conf/icml/Cui0W023}
Cui, J., Huang, W., Wang, Y., and Wang, Y.
\newblock Rethinking weak supervision in helping contrastive learning.
\newblock In \emph{International Conference on Machine Learning (ICML)}, volume 202, pp.\  6448--6467, 2023.

\bibitem[Dao et~al.(2019)Dao, Gu, Ratner, Smith, Sa, and R{\'{e}}]{DBLP:conf/icml/DaoGRSSR19}
Dao, T., Gu, A., Ratner, A., Smith, V., Sa, C.~D., and R{\'{e}}, C.
\newblock A kernel theory of modern data augmentation.
\newblock In \emph{International Conference on Machine Learning (ICML)}, volume~97, pp.\  1528--1537, 2019.

\bibitem[Eckart \& Young(1936)Eckart and Young]{EckartY36}
Eckart, C. and Young, G.
\newblock The approximation of one matrix by another of lower rank.
\newblock \emph{Psychometrika}, 1:\penalty0 211--218, 1936.

\bibitem[Fang et~al.(2023)Fang, Zhang, Hu, Wu, and Xu]{DBLP:journals/tkde/FangZHWX23}
Fang, Q., Zhang, X., Hu, J., Wu, X., and Xu, C.
\newblock Contrastive multi-modal knowledge graph representation learning.
\newblock \emph{{IEEE} Transactions on Knowledge and Data Engineering}, 35\penalty0 (9):\penalty0 8983--8996, 2023.

\bibitem[Ghose et~al.(2023)Ghose, Zhang, Hao, and Coates]{DBLP:conf/aistats/Ghose0HC23}
Ghose, A., Zhang, Y., Hao, J., and Coates, M.
\newblock Spectral augmentations for graph contrastive learning.
\newblock In \emph{International Conference on Artificial Intelligence and Statistics (AISTATS)}, volume 206, pp.\  11213--11266, 2023.

\bibitem[Halko et~al.(2011)Halko, Martinsson, and Tropp]{DBLP:journals/siamrev/HalkoMT11}
Halko, N., Martinsson, P., and Tropp, J.~A.
\newblock Finding structure with randomness: Probabilistic algorithms for constructing approximate matrix decompositions.
\newblock \emph{{SIAM} Review}, 53\penalty0 (2):\penalty0 217--288, 2011.

\bibitem[HaoChen et~al.(2021)HaoChen, Wei, Gaidon, and Ma]{DBLP:conf/nips/HaoChenWGM21}
HaoChen, J.~Z., Wei, C., Gaidon, A., and Ma, T.
\newblock Provable guarantees for self-supervised deep learning with spectral contrastive loss.
\newblock In \emph{Advances in Neural Information Processing Systems (NeurIPS)}, pp.\  5000--5011, 2021.

\bibitem[He et~al.(2020)He, Fan, Wu, Xie, and Girshick]{DBLP:conf/cvpr/He0WXG20}
He, K., Fan, H., Wu, Y., Xie, S., and Girshick, R.~B.
\newblock Momentum contrast for unsupervised visual representation learning.
\newblock In \emph{Conference on Computer Vision and Pattern Recognition (CVPR)}, pp.\  9726--9735, 2020.

\bibitem[Ho et~al.(2020)Ho, Jain, and Abbeel]{DBLP:conf/nips/HoJA20}
Ho, J., Jain, A., and Abbeel, P.
\newblock Denoising diffusion probabilistic models.
\newblock In \emph{Advances in Neural Information Processing Systems (NeurIPS)}, 2020.

\bibitem[Huang et~al.(2023{\natexlab{a}})Huang, Wu, Jie, Zuo, and Zhang]{DBLP:conf/icip/HuangWJZZ23}
Huang, R., Wu, S., Jie, L., Zuo, X., and Zhang, H.
\newblock Siamclim: Text-based pedestrian search via multi-modal siamese contrastive learning.
\newblock In \emph{{IEEE} International Conference on Image Processing (ICIP)}, pp.\  1800--1804, 2023{\natexlab{a}}.

\bibitem[Huang et~al.(2023{\natexlab{b}})Huang, Yi, Zhao, and Jiang]{DBLP:conf/iclr/0001YZJ23}
Huang, W., Yi, M., Zhao, X., and Jiang, Z.
\newblock Towards the generalization of contrastive self-supervised learning.
\newblock In \emph{The Eleventh International Conference on Learning Representations (ICLR)}, 2023{\natexlab{b}}.

\bibitem[Jang \& Wang(2023)Jang and Wang]{DBLP:conf/cvpr/Jang023}
Jang, T. and Wang, X.
\newblock Difficulty-based sampling for debiased contrastive representation learning.
\newblock In \emph{Conference on Computer Vision and Pattern Recognition (CVPR)}, pp.\  24039--24048, 2023.

\bibitem[Jean et~al.(2019)Jean, Wang, Samar, Azzari, Lobell, and Ermon]{DBLP:conf/aaai/JeanWSALE19}
Jean, N., Wang, S., Samar, A., Azzari, G., Lobell, D.~B., and Ermon, S.
\newblock Tile2vec: Unsupervised representation learning for spatially distributed data.
\newblock In \emph{The Thirty-Third {AAAI} Conference on Artificial Intelligence (AAAI)}, pp.\  3967--3974, 2019.

\bibitem[Ji et~al.(2023)Ji, Deng, Nakada, Zou, and Zhang]{DBLP:journals/jmlr/JiDN0Z23}
Ji, W., Deng, Z., Nakada, R., Zou, J., and Zhang, L.
\newblock The power of contrast for feature learning: {A} theoretical analysis.
\newblock \emph{Journal of Machine Learning Research}, 24:\penalty0 330:1--330:78, 2023.

\bibitem[Jing et~al.(2022)Jing, Vincent, LeCun, and Tian]{DBLP:conf/iclr/JingVLT22}
Jing, L., Vincent, P., LeCun, Y., and Tian, Y.
\newblock Understanding dimensional collapse in contrastive self-supervised learning.
\newblock In \emph{International Conference on Learning Representations (ICLR)}, 2022.

\bibitem[Khosla et~al.(2020)Khosla, Teterwak, Wang, Sarna, Tian, Isola, Maschinot, Liu, and Krishnan]{DBLP:conf/nips/KhoslaTWSTIMLK20}
Khosla, P., Teterwak, P., Wang, C., Sarna, A., Tian, Y., Isola, P., Maschinot, A., Liu, C., and Krishnan, D.
\newblock Supervised contrastive learning.
\newblock In \emph{Advances in Neural Information Processing Systems (NeurIPS)}, 2020.

\bibitem[Kilmer et~al.(2021)Kilmer, Horesh, Avron, and Newman]{KilmerHAN21}
Kilmer, M.~E., Horesh, L., Avron, H., and Newman, E.
\newblock Tensor-tensor algebra for optimal representation and compression of multiway data.
\newblock \emph{Proceedings of the National Academy of Sciences}, 118, 2021.

\bibitem[Lee et~al.(2024)Lee, Park, and Lee]{DBLP:conf/iclr/LeePL24}
Lee, S., Park, T., and Lee, K.
\newblock Soft contrastive learning for time series.
\newblock In \emph{International Conference on Learning Representations (ICLR)}, 2024.

\bibitem[Lei et~al.(2023)Lei, Yang, Ying, and Zhou]{DBLP:conf/icml/LeiYYZ23}
Lei, Y., Yang, T., Ying, Y., and Zhou, D.
\newblock Generalization analysis for contrastive representation learning.
\newblock In \emph{International Conference on Machine Learning (ICML)}, volume 202, pp.\  19200--19227, 2023.

\bibitem[Lin et~al.(2023)Lin, Xiao, Dai, Zhang, and Wang]{DBLP:conf/nips/LinXDZW23}
Lin, M., Xiao, T., Dai, E., Zhang, X., and Wang, S.
\newblock Certifiably robust graph contrastive learning.
\newblock In \emph{Advances in Neural Information Processing Systems (NeurIPS)}, 2023.

\bibitem[Lin et~al.(2022)Lin, Zhang, Wang, Shi, Wu, and Zheng]{DBLP:conf/coling/LinZWSW022}
Lin, Z., Zhang, Z., Wang, M., Shi, Y., Wu, X., and Zheng, Y.
\newblock Multi-modal contrastive representation learning for entity alignment.
\newblock In \emph{International Conference on Computational Linguistics (COLING)}, pp.\  2572--2584, 2022.

\bibitem[Liu et~al.(2024)Liu, Tang, and Liu]{DBLP:conf/icml/LiuT024}
Liu, J., Tang, H., and Liu, Y.
\newblock Perfect alignment may be poisonous to graph contrastive learning.
\newblock In \emph{International Conference on Machine Learning (ICML)}, 2024.

\bibitem[Louizos et~al.(2024)Louizos, Reisser, and Korzhenkov]{DBLP:conf/iclr/LouizosRK24}
Louizos, C., Reisser, M., and Korzhenkov, D.
\newblock A mutual information perspective on federated contrastive learning.
\newblock In \emph{International Conference on Learning Representations (ICLR)}, 2024.

\bibitem[Luo et~al.(2023)Luo, Wang, and Wang]{DBLP:conf/iclr/Luo0023}
Luo, R., Wang, Y., and Wang, Y.
\newblock Rethinking the effect of data augmentation in adversarial contrastive learning.
\newblock In \emph{International Conference on Learning Representations (ICLR)}, 2023.

\bibitem[Mikolov et~al.(2013)Mikolov, Sutskever, Chen, Corrado, and Dean]{DBLP:conf/nips/MikolovSCCD13}
Mikolov, T., Sutskever, I., Chen, K., Corrado, G.~S., and Dean, J.
\newblock Distributed representations of words and phrases and their compositionality.
\newblock In \emph{Advances in Neural Information Processing Systems (NIPS)}, pp.\  3111--3119, 2013.

\bibitem[Nonnenmacher et~al.(2022)Nonnenmacher, Oldenburg, Steinwart, and Reeb]{DBLP:conf/icml/NonnenmacherOSR22}
Nonnenmacher, M.~T., Oldenburg, L., Steinwart, I., and Reeb, D.
\newblock Utilizing expert features for contrastive learning of time-series representations.
\newblock In \emph{International Conference on Machine Learning (ICML)}, volume 162, pp.\  16969--16989, 2022.

\bibitem[Pagliardini et~al.(2018)Pagliardini, Gupta, and Jaggi]{DBLP:conf/naacl/PagliardiniGJ18}
Pagliardini, M., Gupta, P., and Jaggi, M.
\newblock Unsupervised learning of sentence embeddings using compositional n-gram features.
\newblock In \emph{Conference of the North American Chapter of the Association for Computational Linguistics: Human Language Technologies (NAACL-HLT)}, pp.\  528--540, 2018.

\bibitem[Rajput et~al.(2019)Rajput, Feng, Charles, Loh, and Papailiopoulos]{DBLP:conf/icml/RajputFCLP19}
Rajput, S., Feng, Z., Charles, Z., Loh, P., and Papailiopoulos, D.~S.
\newblock Does data augmentation lead to positive margin?
\newblock In \emph{International Conference on Machine Learning (ICML)}, volume~97, pp.\  5321--5330, 2019.

\bibitem[Saunshi et~al.(2022)Saunshi, Ash, Goel, Misra, Zhang, Arora, Kakade, and Krishnamurthy]{DBLP:conf/icml/SaunshiAGMZAKK22}
Saunshi, N., Ash, J.~T., Goel, S., Misra, D., Zhang, C., Arora, S., Kakade, S.~M., and Krishnamurthy, A.
\newblock Understanding contrastive learning requires incorporating inductive biases.
\newblock In \emph{International Conference on Machine Learning (ICML)}, volume 162, pp.\  19250--19286, 2022.

\bibitem[Tamkin et~al.(2023)Tamkin, Glasgow, He, and Goodman]{DBLP:conf/nips/TamkinGHG23}
Tamkin, A., Glasgow, M., He, X., and Goodman, N.~D.
\newblock Feature dropout: Revisiting the role of augmentations in contrastive learning.
\newblock In \emph{Advances in Neural Information Processing Systems (NeurIPS))}, 2023.

\bibitem[Tian et~al.(2020)Tian, Sun, Poole, Krishnan, Schmid, and Isola]{DBLP:conf/nips/Tian0PKSI20}
Tian, Y., Sun, C., Poole, B., Krishnan, D., Schmid, C., and Isola, P.
\newblock What makes for good views for contrastive learning?
\newblock In \emph{Advances in Neural Information Processing Systems (NeurIPS)}, 2020.

\bibitem[Tsai et~al.(2022)Tsai, Li, Liu, Liao, Salakhutdinov, and Morency]{DBLP:conf/iclr/TsaiLLLSM22}
Tsai, Y.~H., Li, T., Liu, W., Liao, P., Salakhutdinov, R., and Morency, L.
\newblock Learning weakly-supervised contrastive representations.
\newblock In \emph{International Conference on Learning Representations (ICLR)}, 2022.

\bibitem[van~den Oord et~al.(2018)van~den Oord, Li, and Vinyals]{DBLP:journals/corr/OordLV18}
van~den Oord, A., Li, Y., and Vinyals, O.
\newblock Representation learning with contrastive predictive coding, 2018.

\bibitem[Waida et~al.(2023)Waida, Wada, And{\'{e}}ol, Nakagawa, Zhang, and Kanamori]{DBLP:conf/pkdd/WaidaWANZK23}
Waida, H., Wada, Y., And{\'{e}}ol, L., Nakagawa, T., Zhang, Y., and Kanamori, T.
\newblock Towards understanding the mechanism of contrastive learning via similarity structure: {A} theoretical analysis.
\newblock In \emph{Machine Learning and Knowledge Discovery in Databases: Research Track - European Conference (ECML PKDD)}, volume 14172, pp.\  709--727, 2023.

\bibitem[Wall et~al.(2003)Wall, Rechtsteiner, and Rocha]{Wall2003}
Wall, M.~E., Rechtsteiner, A., and Rocha, L.~M.
\newblock \emph{Singular value decomposition and principal component analysis}, pp.\  91--109.
\newblock Springer US, 2003.

\bibitem[Wang et~al.(2022)Wang, Zhang, Wang, Yang, and Lin]{DBLP:conf/iclr/WangZWYL22}
Wang, Y., Zhang, Q., Wang, Y., Yang, J., and Lin, Z.
\newblock Chaos is a ladder: {A} new theoretical understanding of contrastive learning via augmentation overlap.
\newblock In \emph{International Conference on Learning Representations (ICLR)}, 2022.

\bibitem[Wang et~al.(2024)Wang, Zhang, and Wang]{DBLP:conf/iclr/WangZ024}
Wang, Y., Zhang, J., and Wang, Y.
\newblock Do generated data always help contrastive learning?
\newblock In \emph{International Conference on Learning Representations (ICLR)}, 2024.

\bibitem[Wang et~al.(2023{\natexlab{a}})Wang, Zhao, Cheng, Huang, Liu, Yin, Tang, Li, Wang, Zhang, and Zhao]{DBLP:conf/nips/WangZC23}
Wang, Z., Zhao, Y., Cheng, X., Huang, H., Liu, J., Yin, A., Tang, L., Li, L., Wang, Y., Zhang, Z., and Zhao, Z.
\newblock Connecting multi-modal contrastive representations.
\newblock In \emph{Advances in Neural Information Processing Systems (NeurIPS)}, 2023{\natexlab{a}}.

\bibitem[Wang et~al.(2023{\natexlab{b}})Wang, Zhao, Cheng, Huang, Liu, Yin, Tang, Li, Wang, Zhang, and Zhao]{DBLP:conf/nips/WangZCHLYTLWZZ23}
Wang, Z., Zhao, Y., Cheng, X., Huang, H., Liu, J., Yin, A., Tang, L., Li, L., Wang, Y., Zhang, Z., and Zhao, Z.
\newblock Connecting multi-modal contrastive representations.
\newblock In \emph{Advances in Neural Information Processing Systems (NeurIPS)}, 2023{\natexlab{b}}.

\bibitem[Wen et~al.(2024)Wen, Li, Gong, and Chen]{DBLP:conf/ijcai/WenLGC24}
Wen, W., Li, H., Gong, T., and Chen, H.
\newblock Towards sharper generalization bounds for adversarial contrastive learning.
\newblock In \emph{International Joint Conference on Artificial Intelligence (IJCAI)}, pp.\  5190--5198, 2024.

\bibitem[Wu et~al.(2020)Wu, Zhang, Valiant, and R{\'{e}}]{DBLP:conf/icml/WuZVR20}
Wu, S., Zhang, H.~R., Valiant, G., and R{\'{e}}, C.
\newblock On the generalization effects of linear transformations in data augmentation.
\newblock In \emph{International Conference on Machine Learning (ICML)}, volume 119, pp.\  10410--10420, 2020.

\bibitem[Xia et~al.(2022)Xia, Wu, Wang, Chen, and Li]{DBLP:conf/icml/XiaWWCL22}
Xia, J., Wu, L., Wang, G., Chen, J., and Li, S.~Z.
\newblock Progcl: Rethinking hard negative mining in graph contrastive learning.
\newblock In \emph{International Conference on Machine Learning (ICML)}, volume 162, pp.\  24332--24346, 2022.

\bibitem[Xia et~al.(2023)Xia, Wang, Gao, Yang, and Gao]{DBLP:journals/tip/XiaWGYG23}
Xia, W., Wang, T., Gao, Q., Yang, M., and Gao, X.
\newblock Graph embedding contrastive multi-modal representation learning for clustering.
\newblock \emph{{IEEE} Transactions on Image Processing}, 32:\penalty0 1170--1183, 2023.

\bibitem[Xu et~al.(2024)Xu, Moreno, Wei, Marlin, and Rehg]{DBLP:conf/iclr/XuMWMR24}
Xu, M.~A., Moreno, A., Wei, H., Marlin, B.~M., and Rehg, J.~M.
\newblock {REBAR:} retrieval-based reconstruction for time-series contrastive learning.
\newblock In \emph{The Twelfth International Conference on Learning Representations (ICLR)}, 2024.

\bibitem[Xu et~al.(2023{\natexlab{a}})Xu, Zhang, Liu, Sugiyama, and Kankanhalli]{DBLP:conf/nips/XuZLSK23}
Xu, X., Zhang, J., Liu, F., Sugiyama, M., and Kankanhalli, M.~S.
\newblock Enhancing adversarial contrastive learning via adversarial invariant regularization.
\newblock In \emph{Advances in Neural Information Processing Systems (NeurIPS)}, 2023{\natexlab{a}}.

\bibitem[Xu et~al.(2023{\natexlab{b}})Xu, Zhang, Liu, Sugiyama, and Kankanhalli]{DBLP:conf/nips/XuZLSK23a}
Xu, X., Zhang, J., Liu, F., Sugiyama, M., and Kankanhalli, M.~S.
\newblock Efficient adversarial contrastive learning via robustness-aware coreset selection.
\newblock In \emph{Advances in Neural Information Processing Systems (NeurIPS)}, 2023{\natexlab{b}}.

\bibitem[Yu et~al.(2023{\natexlab{a}})Yu, Liu, Wang, Xu, and Liu]{DBLP:conf/iclr/YuLWXL23}
Yu, Q., Liu, Y., Wang, Y., Xu, K., and Liu, J.
\newblock Multimodal federated learning via contrastive representation ensemble.
\newblock In \emph{International Conference on Learning Representations (ICLR)}, 2023{\natexlab{a}}.

\bibitem[Yu et~al.(2023{\natexlab{b}})Yu, Wang, Zhang, Liu, and Shi]{DBLP:conf/nips/YuWZLS23}
Yu, Y., Wang, X., Zhang, M., Liu, N., and Shi, C.
\newblock Provable training for graph contrastive learning.
\newblock In \emph{Advances in Neural Information Processing Systems (NeurIPS)}, 2023{\natexlab{b}}.

\bibitem[Zang et~al.(2024)Zang, Luo, Wang, Zhang, Wang, Li, and You]{DBLP:conf/icml/Zang00Z0L024}
Zang, Z., Luo, H., Wang, K., Zhang, P., Wang, F., Li, S.~Z., and You, Y.
\newblock Diffaug: Enhance unsupervised contrastive learning with domain-knowledge-free diffusion-based data augmentation.
\newblock In \emph{International Conference on Machine Learning (ICML)}, 2024.

\bibitem[Zhang et~al.(2023{\natexlab{a}})Zhang, Sheng, Cai, Wang, and Chua]{DBLP:conf/nips/ZhangSC0C23}
Zhang, A., Sheng, L., Cai, Z., Wang, X., and Chua, T.
\newblock Empowering collaborative filtering with principled adversarial contrastive loss.
\newblock In \emph{Advances in Neural Information Processing Systems (NeurIPS)}, 2023{\natexlab{a}}.

\bibitem[Zhang et~al.(2023{\natexlab{b}})Zhang, Wang, and Wang]{DBLP:conf/icml/ZhangW023}
Zhang, Q., Wang, Y., and Wang, Y.
\newblock On the generalization of multi-modal contrastive learning.
\newblock In \emph{International Conference on Machine Learning (ICML)}, volume 202, pp.\  41677--41693, 2023{\natexlab{b}}.

\bibitem[Zheng et~al.(2021)Zheng, Wang, You, Qian, Zhang, Wang, and Xu]{DBLP:conf/iccv/Zheng0Y0Z0021}
Zheng, M., Wang, F., You, S., Qian, C., Zhang, C., Wang, X., and Xu, C.
\newblock Weakly supervised contrastive learning.
\newblock In \emph{{IEEE/CVF} International Conference on Computer Vision (ICCV)}, pp.\  10022--10031, 2021.

\bibitem[Zheng et~al.(2024)Zheng, Wang, Cheng, Ma, Chen, Sha, and Luo]{DBLP:conf/iclr/ZhengW0MC0L24}
Zheng, X., Wang, T., Cheng, W., Ma, A., Chen, H., Sha, M., and Luo, D.
\newblock Parametric augmentation for time series contrastive learning.
\newblock In \emph{International Conference on Learning Representations (ICLR)}, 2024.

\bibitem[Zimmermann et~al.(2021)Zimmermann, Sharma, Schneider, Bethge, and Brendel]{DBLP:conf/icml/ZimmermannSSBB21}
Zimmermann, R.~S., Sharma, Y., Schneider, S., Bethge, M., and Brendel, W.
\newblock Contrastive learning inverts the data generating process.
\newblock In \emph{International Conference on Machine Learning (ICML)}, volume 139, pp.\  12979--12990, 2021.

\bibitem[Zou \& Liu(2023)Zou and Liu]{DBLP:journals/jmlr/ZouL23}
Zou, X. and Liu, W.
\newblock Generalization bounds for adversarial contrastive learning.
\newblock \emph{Journal of Machine Learning Research}, 24:\penalty0 114:1--114:54, 2023.

\end{thebibliography}
\bibliographystyle{icml2025}

\newpage
\appendix
\onecolumn
\section{Notations}
The main notations of this paper are summarized in Table \ref{notations}.
\begin{table}[!ht]
    \centering
    \renewcommand\arraystretch{1.5}
    \caption{Summary of main notations involved in this paper.}
    \begin{tabular}{c|l}
        \hline
        Notations & Descriptions\\
        \hline
        $\bar{\mathcal{D}}$ & the original unlabeled dataset drawing from the distribution $\mathcal{P}$\\
        $p(x|\bar{x})$ & the distribution of the augmented sample $x$ conditioned on the unlabeled original sample $\bar{x}$\\
        $p(x, x^+)$ & the joint distribution of the positive augmented sample pair $(x, x^+)$\\
        $\{x_i^-\}_{i=1}^M$ & $M$ negative augmented samples\\
        $\mathcal{F}_1, \mathcal{F}_2$ & the functional spaces of the encoder $f$ and the linear projection head $g$\\
        $k$ & the dimension of embedding vector\\
        $K$ & the number of labels for downstream classification task\\
        $\mathcal{T}$ & the data augmentation set defined as $\{t|t: \mathbb{R}^d \rightarrow \mathbb{R}^d\}$\\
        $\mathcal{L}_{InfoNCE}(f)$ & population InfoNCE loss\\
        $\hat{\mathcal{L}}_{InfoNCE}(f)$ & empirical InfoNCE loss\\
        $f^*$ & the empirical optimal encoder for $\min_{f\in \mathcal{F}_1} \hat{\mathcal{L}}_{InfoNCE}(f)$\\
        $\mathcal{L}_{CE}(g)$ & cross entropy (CE) loss \\
        $g_{f,W}, g_{f,\mu}$ & linear classifier and mean classifier\\
        $W$ & the weight of the linear projection head $g$ defined as $[w_1,...,w_K]$\\
        $\mu$ & the parameter of mean projection head defined as $[\mu_1, ..., \mu_K]$\\
        $\mathcal{E}(f, W)$ & the downstream classification error defined as $\underset{x\sim \mathcal{P}}{\mathrm{Pr}}\left[g_{f,W}(x) \neq y_{x}\right]$\\
        $\alpha$ & the labeling error defined as $\mathbb{E}_{\bar{x}\sim \mathcal{P}, x\sim p(\cdot|\bar{x})}\left[\mathbb{I}\left[y_x\neq y_{\bar{x}}\right]\right]$\\
        $m, m^\prime$ & the size of the matrix $X\in \mathbb{R}^{m\times m^\prime}$ (for example an image)\\
        $U(V)$ & the left $($right$)$ singular matrix with $m (m^\prime)$ orthonormal column vectors\\
        $S$ & the diagonal matrix whose diagonal elements are singular values\\
        $q, q^*$ & the truncated parameter of truncated SVD and its optimal value\\
        $\mathcal{L}_{spe}$ & the spectral contrastive loss\\
        $\mathcal{D}_{aug}, n$ & the augmentation dataset defined as $\{x|x=t(\bar{x}), \bar{x}\in\bar{\mathcal{D}}, t\in \mathcal{T}\}$ and its number\\
        $\mathcal{G}, A, L$ & augmentation graph, its adjacency matrix and normalized graph Laplacian matrix\\
        $\lambda$ & the eigenvalues of $L$\\
        \hline
    \end{tabular}
    \label{notations}
\end{table}

\section{Lemmas}
\begin{lemma}[Lemma A.2 in \citet{DBLP:conf/iclr/WangZWYL22}]
\label{Lemma of the upper bound of the approximation error}
    For $\mathrm{LSE} := \log \mathbb{E}_{p(z)}\left[ \exp \left(f(x)^\top g(z)\right)\right]$, we denote its (biased) Monte Carlo estimate with $M$ random samples $z_i\sim p(z), i=1,...,M$ as $\hat{\mathrm{LSE}}_M = \log \frac{1}{M} \sum_{i=1}^M \exp \left(f(x)^\top g(z_i)\right)$.
    Then, approximation error can be upper bounded in expectation as 
    \begin{align*}
        \mathbb{E}_{x,z_i}\left[\left|\hat{\mathrm{LSE}}_M - \mathrm{LSE}\right|\right] \leq \mathcal{O}\left(M^{-\frac{1}{2}}\right).
    \end{align*}
\end{lemma}

\begin{lemma}[Equation (11) in \citet{DBLP:conf/iclr/WangZWYL22}]
\label{Lemma of the inequality of dot product}
    Let a projector $f\in \mathcal{F}: \mathbb{R}^d \rightarrow \mathbb{S}^{k-1}$ map from the $d$-dimensonal input space to a unit hypersphere in the $k$-dimensional space. 
    For $x, x^+\in \mathbb{R}^d$, we have
    \begin{align*}
        f(x)\left(f(x^+) - \mu_y\right) \leq \left(\frac{f(x^+) - \mu_y}{\left\|f(x^+) - \mu_y \right\|}\right)^\top \left(f(x^+) - \mu_y\right) = \left\|f(x^+) - \mu_y \right\|,
    \end{align*}
    where $\mu_y = \mathbb{E}_{p(x|y)}[f(x)]$, $y$ denotes label.
\end{lemma}

\begin{lemma}[Corollary 3.5 in \citet{Budimir2000FurtherRR}]
\label{Lemma of the reversed Jensen’s inequality}
    Let the function $g: \mathbb{R}^d \rightarrow \mathbb{R}$ be a differentiable convex and $L$-smooth mapping. 
    Then, for any $z\in \mathbb{R}^d$, we have 
    \begin{align*}
        0\leq \mathbb{E}_{p(z)}\left[g(z)\right] - g(\mathbb{E}_{p(z)}\left[z\right]) \leq L \sum_{j=1}^d V(z^{(j)}) = L V(z),
    \end{align*}
    where $z^{(j)}$ denotes the $j$-th  dimension of $z$, $V(z^{(j)})$ denotes the variance of $z^{(j)}$.
\end{lemma}

\begin{lemma}[Theorem 4.5 in \citet{DBLP:conf/icml/AroraKKPS19}]
\label{Lemma of the relationship between real classification risk and mean classification risk}
    With probability at least $1-\delta$, for any $f^*\in \arg\min_{f\in \mathcal{F}_1} \hat{\mathcal{L}}_{InfoNCE}(f)$ and $g\in \mathcal{F}_2$, there holds that $\mathcal{L}_{CE}(g_{f^*,W}) \leq \mathcal{L}_{CE} (g_{f^*,\mu})$.
\end{lemma}

\begin{lemma}[Theorem C.3 in \citet{DBLP:conf/nips/HaoChenWGM21}]
\label{Lemma of downstream classification accuracy}
    Assume the set of augmented data $\mathcal{D}_{aug}$ is finite. 
    Let $f^*\in \arg\min_{f:\mathcal{D}_{aug}\rightarrow \mathbb{R}^k}$ be a minimizer of the population spectral contrastive loss $\mathcal{L}_{spe}(f)$ with $k\in \mathcal{Z}^+$.
    Then, there exists a linear head $W^*\in \mathbb{R}^{k\times K}$ with norm $\|W^*\|_F\leq 1/(1-\lambda_k)$ such that 
    \begin{align*}
        \underset{\bar{x}\sim \mathcal{P}, x\sim p(\cdot|\bar{x})}{\mathrm{Pr}}\left[g_{f^*,W^*}(x) \neq y_{\bar{x}}\right] \leq \frac{4\alpha}{\lambda_{k+1}} + 8\alpha.
    \end{align*}
\end{lemma}

\section{Proofs of main results}
\begin{theorem}[Theorem \ref{Theorem of the bounds of the mean classification risk} (restated)]
    Let Assumption \ref{Assumption of labeling error} hold. 
    For any $f\in \mathcal{F}_1, g\in \mathcal{F}_2$, the gap between the mean downstream classification risk and the contrastive risk $\mathcal{L}_{CE}(g_{f,\mu}) - \mathcal{L}_{InfoNCE}(f)$ can be upper bounded by
    \begin{align*}
        \sqrt{V\left(f(x)\right)} + \sqrt{V^-\left(f(x)\right)} + \mathcal{O}\left(M^{-\frac{1}{2}}\right) - \log\left(\frac{M}{K}\right)
    \end{align*}
    and lower bounded by 
    \begin{align*}
        - \sqrt{V\left(f(x)\right)} - \sqrt{V^-\left(f(x)\right)} - \frac{1}{2} V(f(x^-)) - \mathcal{O}\left(M^{-\frac{1}{2}}\right) - \log\left(\frac{M+1}{K}\right),
    \end{align*}
    where $V\left(f(x)\right) = \mathbb{E}_{(x,x^+)\in X^+}\left[\left\|f(x)-\mu_{y_{x}}\right\|^2\right]$, $V^-\left(f(x)\right) = \mathbb{E}_{(x,x^+)\in X^-} \left[\left\|f(x^+)-\mu_{y_{x}}\right\|^2\right]$, $V(f(x^-)) = V(z|z\in \{f(x^+), f(x^-)\}, y_{x^+}=y_x) = \mathbb{E}_{\{z|z\in \{f(x^+), f(x^-)\}, y_{x^+}=y_x\}} \left[\left\|z - \mu_{y_{z}} \right\|^2\right]$ are the intra-class variance of the representations for true positive augmented samples, the variance for false positive augmented samples, and the intra-class variance for negative and true positive augmented samples, respectively. 
\end{theorem}
\begin{theorem1proof}
\label{theorem1proof}
Assume positive augmented samples are $x, x^+$ and $M$ negative samples are $x_i^-, i=1,...,M$ ($x_i^-$ belongs to any class in the $K$ classes).
We let $\mu_{y_x} = \mathbb{E}_{\{x|y=y_x\}}\left[f(x)\right]$. 
Under our assumptions, the labels of $x$ and $x^+$ may be different from each other. 
In other words, data augmentation may change the semantics of the original samples due to the intrinsic randomness. 
Thus, we denote the sets of true positive sample pair and false positive sample pair as $X^+ = \left\{(x,x^+)| y_x = y_{x^+}\right\}$ and $X^- = \left\{(x,x^+)| y_x \neq y_{x^+}\right\}$, respectively. 
Then, we have the following lower bounds of the InfoNCE loss
\begin{align*}
    & \mathcal{L}_{InfoNCE}(f)\\
    = & \mathbb{E}_{x,x^+, \{x_i^-\}_{i=1}^M} \left[-\log \frac{e^{f(x)^\top f(x^+)}}{e^{f(x)^\top f(x^+)} + \sum_{i=1}^M e^{f(x)^\top f(x_i^-)}}\right]\\
    = & \mathbb{E}_{x,x^+, \{x_i^-\}_{i=1}^M} \left[\log\left(1 +  \frac{\sum_{i=1}^M \exp\left(f(x)^\top f(x_i^-)\right)}{\exp\left(f(x)^\top f(x^+)\right)}\right)\right]\\
    \geq & -\mathbb{E}_{x,x^+}\left[f(x)^\top f(x^+)\right] + \mathbb{E}_{x, \{x_i^-\}_{i=1}^M}\left[\log \sum_{i=1}^M \exp \left(f(x)^\top f(x_i^-)\right)\right]\\
    = & -\mathbb{E}_{x,x^+}\left[f(x)^\top f(x^+)\right] + \mathbb{E}_{x, \{x_i^-\}_{i=1}^M}\left[\log \frac{1}{M} \sum_{i=1}^M \exp \left(f(x)^\top f(x_i^-)\right)\right] + \log M\\
    \overset{(1)}{\geq} & -\mathbb{E}_{x,x^+}\left[f(x)^\top f(x^+)\right] + \mathbb{E}_{x}\left[\log \frac{1}{M} \mathbb{E}_{\{x_i^-\}_{i=1}^M}\left[\sum_{i=1}^M \exp \left(f(x)^\top f(x_i^-)\right)\right]\right] - \mathcal{O}\left(M^{-\frac{1}{2}}\right) + \log M\\
    = & -\mathbb{E}_{x,x^+}\left[f(x)^\top f(x^+)\right] + \mathbb{E}_{x}\left[\log \mathbb{E}_{x^-}\left[\exp \left(f(x)^\top f(x^-)\right)\right]\right] - \mathcal{O}\left(M^{-\frac{1}{2}}\right) + \log M\\
    = & -\mathbb{E}_{(x,x^+)\in X^+}\left[f(x)^\top f(x^+)\right] - \mathbb{E}_{(x,x^+)\in X^-}\left[f(x)^\top f(x^+)\right]\\
    & + \mathbb{E}_{x}\left[\log \mathbb{E}_{y^-} \left[\mathbb{E}_{\{x^-|y_{x^-} = y^-\}}\left[\exp \left(f(x)^\top f(x^-)\right)\right]\right]\right] - \mathcal{O}\left(M^{-\frac{1}{2}}\right) + \log M\\
    \overset{(2)}{\geq} & -\mathbb{E}_{(x,x^+)\in X^+}\left[f(x)^\top \left(\mu_{y_{x}} + f(x^+) - \mu_{y_{x}}\right)\right]  - \mathbb{E}_{(x,x^+)\in X^-}\left[f(x)^\top \left(\mu_{y_x} + f(x^+) - \mu_{y_x}\right)\right]\\
    & + \mathbb{E}_{x}\left[\log \mathbb{E}_{y^-}\left[\exp \left(f(x)^\top \mu_{y^-} \right)\right]\right] - \mathcal{O}\left(M^{-\frac{1}{2}}\right) + \log M\\
    = & -\mathbb{E}_{(x,x^+)\in X^+}\left[f(x)^\top \mu_{y_{x}} + f(x)^\top \left(f(x^+) - \mu_{y_{x}}\right)\right] - \mathbb{E}_{(x,x^+)\in X^-} \left[f(x)^\top \mu_{y_{x}} + f(x)^\top \left(f(x^+) - \mu_{y_{x}}\right)\right]\\
    & + \mathbb{E}_{x}\left[\log \mathbb{E}_{y^-}\left[\exp \left(f(x)^\top \mu_{y^-} \right)\right]\right] - \mathcal{O}\left(M^{-\frac{1}{2}}\right) + \log M\\
    \overset{(3)}{\geq} & -\mathbb{E}_{(x,x^+)\in X^+}\left[f(x)^\top \mu_{y_{x}} + \left\|f(x^+) - \mu_{y_{x}}\right\|\right] + \mathbb{E}_{x}\left[\log \mathbb{E}_{y^-}\left[\exp \left(f(x)^\top \mu_{y^-} \right)\right]\right]\\
    & - \mathcal{O}\left(M^{-\frac{1}{2}}\right) + \log M - \mathbb{E}_{(x,x^+) \in X^-}\left[f(x)^\top \mu_{y_{x}} + \left\|f(x^+) - \mu_{y_{x}}\right\|\right]\\
    \overset{(4)}{\geq} & -\mathbb{E}_{(x,x^+)\in X^+} \left[f(x)^\top \mu_{y_{x}}\right] - \sqrt{\mathbb{E}_{(x,x^+)\in X^+}\left[\left\|f(x) - \mu_{y_{x}}\right\|^2\right]} + \mathbb{E}_{x}\left[\log \frac{1}{K} \sum_{k=1}^K \exp \left(f(x)^\top \mu_{k} \right)\right]\\
    & - \mathcal{O}\left(M^{-\frac{1}{2}}\right) + \log M - \mathbb{E}_{(x,x^+)\in X^-}\left[f(x)^\top \mu_{y_{x}}\right] - \sqrt{\mathbb{E}_{(x,x^+)\in X^-}\left[\left\|f(x^+) - \mu_{y_{x}}\right\|^2\right]}\\
    = & -\mathbb{E}_{x}\left[f(x)^\top \mu_{y_{x}} - \log \sum_{k=1}^K \exp \left(f(x)^\top \mu_{k} \right)\right] - \sqrt{\mathbb{E}_{(x,x^+)\in X^+}\left[\left\|f(x) - \mu_{y_{x}}\right\|^2\right]}\\
    & - \mathcal{O}\left(M^{-\frac{1}{2}}\right) + \log \left(\frac{M}{K}\right) - \sqrt{\mathbb{E}_{(x,x^+)\in X^-}\left[\left\|f(x^+) - \mu_{y_{x}}\right\|^2\right]}\\
    = & \mathcal{L}_{CE}(g_{f,\mu}) - \sqrt{\mathbb{E}_{(x,x^+)\in X^+}\left[\left\|f(x) - \mu_{y_{x}}\right\|^2\right]} - \mathcal{O}\left(M^{-\frac{1}{2}}\right) + \log \left(\frac{M}{K}\right)  - \sqrt{\mathbb{E}_{(x,x^+)\in X^-}\left[\left\|f(x^+) - \mu_{y_{x}}\right\|^2\right]},
\end{align*}
where inequality (1) derives from Lemma \ref{Lemma of the upper bound of the approximation error}, inequality (2) is due to the Jensen’s inequality of convex function (exponential function $\exp$), inequality (3) follows Lemma \ref{Lemma of the inequality of dot product}, and inequality (4) is from the Cauchy–Schwarz inequality.
Let $V\left(f(x)\right)=\mathbb{E}_{(x,x^+)\in X^+}\left[\left\|f(x) - \mu_{y_{x}}\right\|^2\right]$ and $V^-\left(f(x)\right)=\mathbb{E}_{(x,x^+)\in X^-}\left[\left\|f(x^+) - \mu_{y_{x}}\right\|^2\right]$.
We can get the upper bound of $\mathcal{L}_{CE}(g_{f,\mu}) + \log\left(\frac{M}{K}\right) - \mathcal{L}_{InfoNCE}(f)$ as
\begin{align*}
    \mathcal{L}_{CE}(g_{f,\mu}) + \log\left(\frac{M}{K}\right) - \mathcal{L}_{InfoNCE}(f) 
    \leq \sqrt{V^-\left(f(x)\right)} + \sqrt{V\left(f(x)\right)} + \mathcal{O}\left(M^{-\frac{1}{2}}\right).
\end{align*}
Next, we will prove the corresponding lower bound. 
\begin{align*}
    & \mathcal{L}_{CE}(g_{f,\mu})\\
    = & -\mathbb{E}_{x}\left[f(x)^\top \mu_{y_{x}}\right] + \mathbb{E}_{x}\left[\log \sum_{k=1}^K \exp\left(f(x)^\top \mu_k\right)\right]\\
    = & -\mathbb{E}_{x}\left[f(x)^\top \mu_{y_{x}}\right] + \mathbb{E}_{x}\left[\log \frac{1}{K} \sum_{k=1}^K \exp\left(f(x)^\top \mu_k\right)\right] + \log K\\
    = & -\mathbb{E}_{(x,x^+)\in X^+}\left[f(x)^\top \mu_{y_{x}}\right] + \mathbb{E}_{x}\left[\log \mathbb{E}_{y^-} \left[\exp\left(f(x)^\top \mu_{y^-}\right) \right]\right] + \log K - \mathbb{E}_{(x,x^+)\in X^-} \left[f(x)^\top \mu_{y_{x}}\right]\\
    \overset{(1)}{\geq} & -\mathbb{E}_{(x,x^+)\in X^+}\left[f(x)^\top f(x^+) + f(x)^\top \left(\mu_{y_{x}} - f(x^+) \right)\right]\\
    & + \mathbb{E}_{x}\left[\mathbb{E}_{y_{x},\{y_i^-\}_{i=1}^M} \left[\log \frac{1}{M+1} \left(\exp\left(f(x)^\top \mu_{y_{x}}\right) + \sum_{i=1}^M \exp\left(f(x)^\top \mu_{y_i^-}\right)\right) \right]\right]\\
    & - \mathcal{O}\left(M^{-\frac{1}{2}}\right) + \log K - \mathbb{E}_{(x,x^+)\in X^-} \left[f(x)^\top f(x^+) + f(x)^\top \left(\mu_{y_{x}} - f(x^+)\right)\right]\\
    \overset{(2)}{\geq} & -\mathbb{E}_{(x,x^+)\in X^+} \left[f(x)^\top f(x^+)\right] -\mathbb{E}_{(x,x^+)\in X^+} \left[\left\|f(x)-\mu_{y_{x}}\right\|\right] \\
    & + \mathbb{E}_{x}\left[\mathbb{E}_{y_{x},\{y_i^-\}_{i=1}^M} \left[\log \frac{1}{M+1} \left(\exp\left(f(x)^\top \mu_{y_{x}}\right) + \sum_{i=1}^M \exp\left(f(x)^\top \mu_{y_i^-}\right)\right) \right]\right]\\
    & - \mathcal{O}\left(M^{-\frac{1}{2}}\right) + \log K - \mathbb{E}_{(x,x^+)\in X^-} \left[f(x)^\top f(x^+)\right] - \mathbb{E}_{(x,x^+)\in X^-} \left[\|f(x^+) - \mu_{y_{x}}\|\right]\\
    = & -\mathbb{E}_{x,x^+}\left[f(x)^\top f(x^+)\right] + \mathbb{E}_{x}\left[\mathbb{E}_{y_{x},\{y_i^-\}_{i=1}^M} \left[\log \frac{1}{M+1} \left(\exp\left(f(x)^\top \mu_{y_{x}}\right) + \sum_{i=1}^M \exp\left(f(x)^\top \mu_{y_i^-}\right)\right) \right]\right]\\
    & - \mathbb{E}_{(x,x^+)\in X^+} \left[\left\|f(x)-\mu_{y_{x}}\right\|\right] - \mathbb{E}_{(x,x^+)\in X^-} \left[\left\|f(x^+)-\mu_{y_{x}}\right\|\right] - \mathcal{O}\left(M^{-\frac{1}{2}}\right) + \log K\\
    = & -\mathbb{E}_{x,x^+}\left[f(x)^\top f(x^+)\right] - \mathbb{E}_{(x,x^+)\in X^+} \left[\left\|f(x)-\mu_{y_{x}}\right\|\right] - \mathbb{E}_{(x,x^+)\in X^-} \left[\left\|f(x^+)-\mu_{y_{x}}\right\|\right] - \mathcal{O}\left(M^{-\frac{1}{2}}\right) + \log K\\
    & + \mathbb{E}_{x}\left[\mathbb{E}_{y_{x},\{y_i^-\}_{i=1}^M} \left[\log \frac{1}{M+1} \left(\mathbb{E}_{\left\{x^+, x_i^-| y_{x^+} = y_{x}, y_{x_i^-} = y_i^-\right\}} \left[\exp\left(f(x)^\top f(x^+)\right) + \sum_{i=1}^M \exp\left(f(x)^\top f(x_i^-)\right)\right]\right) \right]\right]\\
    \overset{(3)}{\geq} & -\mathbb{E}_{x,x^+}\left[f(x)^\top f(x^+)\right] - \sqrt{\mathbb{E}_{(x,x^+)\in X^+}\left[\left\|f(x)-\mu_{y_{x}}\right\|^2\right]} - \sqrt{\mathbb{E}_{(x,x^+)\in X^-}\left[\left\|f(x^+) -\mu_{y_{x}}\right\|^2\right]} - \mathcal{O}\left(M^{-\frac{1}{2}}\right) + \log K\\
    & + \mathbb{E}_{x}\left[\mathbb{E}_{y_{\bar{x}},\{y_i^-\}_{i=1}^M} \left[\mathbb{E}_{\left\{x^+, x_i^-| y_{x^+} = y_{x}, y_{x_i^-} = y_i^-\right\}}\left[\log \frac{1}{M+1} \left(\exp\left(f(x)^\top f(x^+)\right) + \sum_{i=1}^M \exp\left(f(x)^\top f(x_i^-)\right)\right)\right] \right]\right]\\
    & - \frac{1}{2} V(z|z\in \{f(x^+), f(x^-)\}, y_{x^+}=y_x)\\
    = & -\mathbb{E}_{x,x^+}\left[f(x)^\top f(x^+)\right] + \mathbb{E}_{x}\left[\mathbb{E}_{x^+,\{x_i^-\}_{i=1}^M} \left[\log \left(\exp\left(f(x)^\top f(x^+)\right) + \sum_{i=1}^M \exp\left(f(x)^\top f(x_i^-)\right)\right)\right] \right]\\
    & - \sqrt{V(f(x))} - \sqrt{V^-(f(x))} - \frac{1}{2} V(z|z\in \{f(x^+), f(x^-)\}, y_{x^+}=y_x) - \mathcal{O}\left(M^{-\frac{1}{2}}\right) - \log \left(\frac{M+1}{K}\right)\\
    = & \mathcal{L}_{InfoNCE}(f) - \sqrt{V(f(x))} - \sqrt{V^-(f(x))} - \frac{1}{2} V(z|z\in \{f(x^+), f(x^-)\}, y_{x^+}=y_x) - \mathcal{O}\left(M^{-\frac{1}{2}}\right) - \log \left(\frac{M+1}{K}\right),
\end{align*}
where the inequalities (1), (2) are similar to the proof of the above upper bound, the inequality (3) follows Lemma \ref{Lemma of the reversed Jensen’s inequality} 
\footnote{\citet{DBLP:conf/iclr/WangZWYL22} proved the convex function logsumexp is $\frac{1}{2}$-smooth.} 
and the Cauchy–Schwarz inequality.
We can get the lower bound of $\mathcal{L}_{CE}(g_{f,\mu}) + \log\left(\frac{M+1}{K}\right) - \mathcal{L}_{InfoNCE}(f)$ as 
\begin{align*}
    & \mathcal{L}_{CE}(g_{f,\mu}) + \log\left(\frac{M+1}{K}\right) -\mathcal{L}_{InfoNCE}(f)\\
    \geq & - \sqrt{V\left(f(x)\right)} - \sqrt{V^-\left(f(x)\right)} - \frac{1}{2} V(z|z\in \{f(x^+), f(x^-)\}, y_{x^+}=y_x) - \mathcal{O}\left(M^{-\frac{1}{2}}\right).
\end{align*}
\end{theorem1proof}

By integrating the Theorem 4.5 of \citet{DBLP:conf/icml/AroraKKPS19}, we can directly derive Corollary \ref{Corollary of the upper bound of the classification risk} which measures the upper bound of the classification risk $\mathcal{L}_{CE}(g_{f^*,W})$.

\begin{corollary}
\label{Corollary of the upper bound of the classification risk}
    Under the condition of Theorem \ref{Theorem of the bounds of the mean classification risk}, the downstream classification risk $\mathcal{L}_{CE}(g_{f^*,W}) + \log \left(\frac{M}{K}\right)$ can be upper bounded by 
        $\mathcal{L}_{InfoNCE}(f^*) + \sqrt{V^-\left(f^*(x)\right)} + \sqrt{V\left(f^*(x)\right)} + \mathcal{O}\left(M^{-\frac{1}{2}}\right).$
\end{corollary}


\begin{corollary1proof}
\label{corollary1proof}
According to Lemma \ref{Lemma of the relationship between real classification risk and mean classification risk}, we can know that, for any $f^*\in \arg\min_{f\in\mathcal{F}_1} \hat{\mathcal{L}}_{InfoNCE}(f)$ and $g\in \mathcal{F}_2$, the downstream classification risk $\mathcal{L}_{CE}(g_{f^*,W})$ with the linear classifier $W$ can be bounded by the mean downstream classification risk $\mathcal{L}_{CE}(g_{f^*,\mu})$ with the mean classifier $\mu$. 
And, from Theorem \ref{Theorem of the bounds of the mean classification risk}, we can obtain the upper bound of the mean downstream classification risk $\mathcal{L}_{CE}(g_{f^*,\mu})$ as 
$\mathcal{L}_{InfoNCE}(f^*) + \sqrt{V^-\left(f^*(x)\right)} + \sqrt{V\left(f^*(x)\right)} + \mathcal{O}\left(M^{-\frac{1}{2}}\right) - \log \left(\frac{M}{K}\right)$.
Therefore, the upper bound of Corollary \ref{Corollary of the upper bound of the classification risk} is proved.
\end{corollary1proof}

\begin{assumption}[Assumption \ref{Proposition of the effect of SVD on the labeling error} (restated)]
    Let a sample and its corresponding sample after applying SVD be represented as the matrices $X$ and $\hat{X}_q\in \mathbb{R}^{m\times m^\prime}$, respectively. 
    Assume that there are $q^*$ singular values associated with semantic-related information.
    When $q\geq q^*$, under Assumption \ref{Assumption of labeling error} and the augmentation collection $\mathcal{T}$, the true label of the augmented sample of $\hat{X}_q$ is inconsistent with the latent label of $X$ with the probability $\alpha_q \leq \alpha$.
    Conversely, when $q<q^*$, the corresponding probability satisfies $\alpha_q > \alpha_{q^*}$.
\end{assumption}

\begin{proposition1proof}
In practice, each sample possesses a unique amount of semantic-related information content. 
While for convenience, we assume that each sample can be decomposed into $q^*$ singular values capturing semantic-related information.
Under Lemma \ref{Lemma of Eckart-Young Theorem} and the default assumption that there is a positive correlation between the amount of information and the importance of information, we can demonstrate that the larger the singular value, the more semantic-related the information captured in the corresponding subspace. 
Table \ref{Table of any two singular values discarded} show the trend of increase of downstream classification accuracy from the experiment discarding the singular values $s_1, s_2$ to the one discarding $s_{31}, s_{32}$, which empirically validates this demonstration. 
Based on the above analysis, we can derive that truncated SVD firstly discards semantically irrelevant information that leads to labeling error. 
When $q$ takes the value $q^*$, samples don't have any semantically irrelevant information. 
When $q < q^*$, labeling error will increase due to the loss of semantic-related information.

\end{proposition1proof}



\begin{theorem}[Theorem \ref{Theorem of the classification risk bounds on the optimal function after optimal truncated SVD} (restated)]
    Given the conditions of Theorem \ref{Theorem of the bounds of the mean classification risk} and Assumption \ref{Assumption of perfect alignment for f^*}, after taking the truncated SVD on $\bar{\mathcal{D}}$, the mean downstream classification risk $\mathcal{L}_{CE}(g_{f,\mu}) - \mathcal{L}_{InfoNCE}(f)$ with the encoder $f$ can be upper bounded by 
    \begin{align*}
         \epsilon_{q^*} + \epsilon_q + \mathcal{O}\left(M^{-\frac{1}{2}}\right) - \log\left(\frac{M}{K}\right)
    \end{align*}
    and lower bounded by 
    \begin{align*}
        - \epsilon_{q^*} - \epsilon_{q} - \frac{1}{2} V(f(x^-)) - \mathcal{O}\left(M^{-\frac{1}{2}}\right) - \log\left(\frac{M+1}{K}\right).
    \end{align*}
    When $q=q^*$, the two bounds are $\epsilon_{q^*} + \mathcal{O}\left(M^{-\frac{1}{2}}\right) - \log\left(\frac{M}{K}\right)$ and $- \epsilon_{q^*} - V(f(x^-)) - \mathcal{O}\left(M^{-\frac{1}{2}}\right) - \log\left(\frac{M+1}{K}\right)$.
\end{theorem}

\begin{corollary2proof}

\textbf{Upper bound:}
From the proof of Theorem \ref{Theorem of the bounds of the mean classification risk}, we can get that
\begin{align*}
    & \mathcal{L}_{InfoNCE}(f)\\
    \geq & -\mathbb{E}_{(x,x^+)\in X^+}\left[f(x)^\top \mu_{y_{x}} + \left\|f(x^+) - \mu_{y_{x}}\right\|\right] + \mathbb{E}_{x}\left[\log \mathbb{E}_{y^-}\left[\exp \left(f(x)^\top \mu_{y^-} \right)\right]\right]\\
    & - \mathcal{O}\left(M^{-\frac{1}{2}}\right) + \log M - \mathbb{E}_{(x,x^+)\in X^-} \left[f(x)^\top \mu_{y_{x}} + \left\|f(x^+) - \mu_{y_{x}}\right\|\right]\\
    \geq & \mathcal{L}_{CE}(g_{f,\mu}) - \mathbb{E}_{(x,x^+)\in X^+} \left[\left\|f(x) - \mu_{y_{x}}\right\|\right] - \mathcal{O}\left(M^{-\frac{1}{2}}\right) + \log \left(\frac{M}{K}\right) - \mathbb{E}_{(x,x^+)\in X^-} \left[\left\|f(x^+) - \mu_{y_{x}}\right\|\right]\\
    = & \mathcal{L}_{CE}(g_{f,\mu}) - \mathbb{E}_{(x,x^+)\in X^+} \left[\left\|\mathbb{E}_{\{x^+|y_{x^+}=y_{x}\}}[f(x) - f(x^+)]\right\|\right] - \mathcal{O}\left(M^{-\frac{1}{2}}\right) + \log \left(\frac{M}{K}\right)\\
    & - \mathbb{E}_{(x,x^+)\in X^-} \left[\left\|\mathbb{E}_{x}\left[f(x^+) - f(x)\right] \right\|\right]\\
    \overset{(1)}{\geq} & \mathcal{L}_{CE}(g_{f,\mu}) - \mathbb{E}_{(x,x^+)\in X^+} \left[\left\|f(x) - f(x^+)\right\|\right] - \mathcal{O}\left(M^{-\frac{1}{2}}\right) + \log \left(\frac{M}{K}\right) - \mathbb{E}_{(x,x^+)\in X^-} \left[\left\|f(x^+) - f(x)\right\|\right]\\
    \overset{(2)}{\geq} & \mathcal{L}_{CE}(g_{f,\mu}) - \epsilon_{q^*} - \epsilon_{q} - \mathcal{O}\left(M^{-\frac{1}{2}}\right) + \log \left(\frac{M}{K}\right),
\end{align*}
where the inequality (1) is derived from the Cauchy–Schwarz inequality and the inequality (2) is due to Assumption \ref{Assumption of perfect alignment for f^*}. 

Then, 
\begin{align*}
    \mathcal{L}_{CE}(g_{f,\mu}) + \log\left(\frac{M}{K}\right) - \mathcal{L}_{InfoNCE}(f)
    \leq \epsilon_{q^*} + \epsilon_q + \mathcal{O}\left(M^{-\frac{1}{2}}\right).
\end{align*}

\textbf{Lower bound:}
Similarly, we can get that
\begin{align*}
    & \mathcal{L}_{CE}(g_{f,\mu})\\
    \geq & \mathcal{L}_{InfoNCE}(f) - \mathbb{E}_{(x,x^+)\in X^+} \left[\left\|f(x) - \mu_{y_{x}}\right\|\right] - \mathbb{E}_{(x,x^+)\in X^-} \left[\left\|f(x^+) - \mu_{y_{x}}\right\|\right]\\
    & - \frac{1}{2} V(z|z\in \{f(x^+), f(x^-)\}, y_{x^+}=y_x) - \mathcal{O}\left(M^{-\frac{1}{2}}\right) - \log \left(\frac{M+1}{K}\right)\\
    \geq & \mathcal{L}_{InfoNCE}(f) - \epsilon_{q^*}^2 -\epsilon_{q}^2 - \frac{1}{2} V(z|z\in \{f(x^+), f(x^-)\}, y_{x^+}=y_x) - \mathcal{O}\left(M^{-\frac{1}{2}}\right) - \log \left(\frac{M+1}{K}\right).
\end{align*}
\end{corollary2proof}

\begin{corollary}
\label{Corollary of the upper bound of the classification risk after SVD}
    Under the condition of Theorem \ref{Theorem of the classification risk bounds on the optimal function after optimal truncated SVD}, the downstream classification risk $\mathcal{L}_{CE}(g_{f^*,W}) + \log\left(\frac{M}{K}\right)$ can be upper bounded by 
        $\mathcal{L}_{InfoNCE}(f^*) + \epsilon_{q^*} + \epsilon_q + \mathcal{O}\left(M^{-\frac{1}{2}}\right).$
    When $q=q^*$, the bound is $\mathcal{L}_{InfoNCE}(f^*) + \epsilon_{q^*} + \mathcal{O}\left(M^{-\frac{1}{2}}\right)$.
\end{corollary}

\begin{theorem}[Theorem \ref{Theorem of the upper bound of downstream classification error} (restated)]
    Let Assumption \ref{Assumption of labeling error} hold. 
    For the empirical optimal encoder $f^*$ taking the truncated SVD with hyper-parameter $q$ on unlabeled original dataset, there exists a linear head $W^*\in \mathbb{R}^{k\times K}$ with norm $\|W^*\|_F\leq 1/(1-\lambda_{k,q})$ such that 
    \begin{eqnarray*}
        \mathcal{E}(f^*,W^*) \leq \frac{4\alpha_q}{\lambda_{k+1,q}} + 8\alpha_q,
    \end{eqnarray*}
    where $k$ denotes the dimension of embedding space and $\lambda_{k+1, q}$ denotes the $k+1$-th eigenvalues of $L_q$.
\end{theorem}

\begin{corollary3proof}
Lemma \ref{Lemma of downstream classification accuracy} states that 
\begin{align*}
    \underset{\bar{x}\sim \mathcal{P}, x\sim p(\cdot|\bar{x})}{\mathrm{Pr}}\left[g_{f^*,W^*}(x) \neq y_{\bar{x}}\right] \leq \frac{4\alpha}{\lambda_{k+1}} + 8\alpha.
\end{align*}
We rewrite the definition $\mathcal{E}(f, W) = \underset{\bar{x}\sim \mathcal{P}}{\mathrm{Pr}}\left[g_{f,W}(\bar{x}) \neq y_{\bar{x}}\right]$ as $\underset{\bar{x}\sim \mathcal{P}, x\sim p(\bar{x}|\bar{x})}{\mathrm{Pr}}\left[g_{f,W}(x) \neq y_{\bar{x}}\right]$. 
Thus, after taking the truncated SVD with hyper-parameter $q\in[m]$ on unlabeled sample, the following bound holds
\begin{align*}
    \mathcal{E}(f^*, W^*)
    = \underset{\bar{x}\sim \mathcal{P}, x\sim p(\bar{x}|\bar{x})}{\mathrm{Pr}}\left[g_{f^*,W^*}(x) \neq y_{\bar{x}}\right]
    \leq \underset{\bar{x}\sim \mathcal{P}, x\sim p(\cdot|\bar{x})}{\mathrm{Pr}}\left[g_{f^*,W^*}(x) \neq y_{\bar{x}}\right]
    \leq \frac{4\alpha_q}{\lambda_{k+1,q}} + 8\alpha_q.
    \end{align*}
\end{corollary3proof}

\section{Experimental setting}
\label{Experimental setting}
\textbf{Details of models and datasets.} 
For the backbone structure, we use three variants of Resnet, i.e., Resnet-18, Resnet-50, and Resnet-152, where the dimensions of embedding are chosen from 128, 256, 512, 1024, and 2048. 
We use two-layers multilayer perceptron (MLP) to be the projection layers. 
For the datasets, we employ three benchmark datasets, i.e., CIFAR-10, CIFAR-100, and STL-10.

\textbf{Details of pre-training and fine-tuning.}
We do pre-training for 100 epochs with batch size 256 using the optimizer Adam. 
The optimizer has the weight decay parameter 0.0004 and learning rate 0.0003. 
The learning rate decreases following the cosine schedule. 
In this stage, the dimension of the final output of the model is set as 128.
For the downstream fine-tuning process, we train with a small number of labeling samples for 100 epochs with batch size 256 using the optimizer Adam with weight decay parameter 0.0008 and learning rate 0.0003. 
We use 1 RTX 2070 GPU for all experiments.

\textbf{Details of these experiments with inflation}
Considering fair comparison, we do pre-training for 33 epochs with batch size 256 when we make data inflation. 
The settings of fine-tuning are the same for all experiments.

\textbf{Details of augmentations}
In the main text, we provide several groups of augmentation strategies whose components are listed. 
We need to make the explanations that these abbreviations of augmentations without specific parameter settings adopt their default settings.
For example, “RRC” (instead of “RRC(0.08, 0.5)”) adopts the default parameters “(0.08, 1.0)”, where the range “(0.08, 1.0)” denotes the proportion of preserved area after cropping. 
“Color jitter” (instead of “Color jitter(0.5, 0.4)”) adopts the default parameters “(1.0, 0.8)”, where “1.0” denotes a scale parameter and “0.8” denotes a probability parameter. 
“Cutout” (instead of “Cutout(0.5, 1.0)”) adopts the default parameters “(0.1, 1.0)”, where the range “(0.1, 1.0)” denotes the coordinate parameters used to cut samples.

\section{Discussions with related works}
\citet{DBLP:conf/iclr/0001YZJ23} proposed the concept of augmented distance and provided some upper bounds revealing the theoretical effect of augmented distance on understream classification performance. 
Specifically, they found that the classification performance of contrastive SSL is related to three key factors: alignment of positive samples, divergence of class centers, and concentration of augmented data. 
Theorem 2 in our work provided both upper and lower bound, which can not only give similar conclusions but also reveal some additional factors. 
Firstly, the term $V(f(x)|y_{\bar{x}})$ implies the alignment of positive samples. 
Secondly, the term $V_{y_{\bar{x}}^\neg}(f(x)|y_{\bar{x}})$ stems from labeling error caused by data augmentation, which is similar to the concentration of augmented data. 
Thirdly, the term $V(f(x^-)|y^-)$ implies the alignment of negative samples, which is not considered by \citet{DBLP:conf/iclr/0001YZJ23}. 
More importantly, we further improved the bounds of Theorem 2 via data dimensionality reduction and provided the corresponding theoretical analysis and empirical observations (Section 4.2).

\citet{DBLP:conf/icml/Cui0W023} established a theoretical framework for weakly supervised contrastive learning for the first time. 
Their results revealed that 1) semi-supervised information improves the error bound compared with purely unsupervised contrastive learning by using all labeled samples; 2) joint training of supervised and unsupervised contrastive learning does not improve the error bound compared with purely supervised or purely unsupervised contrastive learning. 
Although weakly supervised contrastive learning is not the topic of our work, the labeling error considered by our work is analogous to a type of weak supervision, i.e., noisy-labeled information. 
Therefore, we will extend the theoretical analysis of this work to weakly supervised contrastive learning in our future work. 
Besides, \citet{DBLP:conf/icml/Cui0W023} and our work both gave the suggestion that we should choose a moderate feature dimension $k$, which enhances the credibility of our suggestion.

\section{Limitations}
\label{Limitations}
The limitations of our work are listed as follows:
\begin{itemize}
    \item This paper selects the embedding dimension $k=512$ or $k=1024$ as an example of a moderate choice. 
    Nevertheless, we've observed that the optimal values of $k$ vary somewhat across different experimental settings, showing a degree of inconsistency. 
    Delving into the underlying reasons for this inconsistency holds great interest since it has the potential to provide valuable insights that would be beneficial for the design of more effective algorithms.

    \item This paper pioneers in theoretically elucidating the role of labeling error in contrastive learning from a novel perspective, data dimensionality reduction. 
    Algorithmically, we utilize the classical SVD for preliminary empirical validation. 
    However, considering the unavailability and diversity of $p^*$ across different settings, the fixed $q$ in the traditional SVD fails to guarantee a sufficiently small labeling error. 
    Thus, our future research will focus on developing a more flexible low-rank image approximation method, which assigns a distinct truncated parameter $q$ to each sample.

    \item Even though our work mainly concentrates on the study of the potential false positive augmented samples, conducting research on false negative augmented samples remains of great necessity.
\end{itemize}

\section{Other Experimental Results}
\label{Other Experimental Results}

\begin{figure*}[ht]
    \centering
    \includegraphics[width=\textwidth]{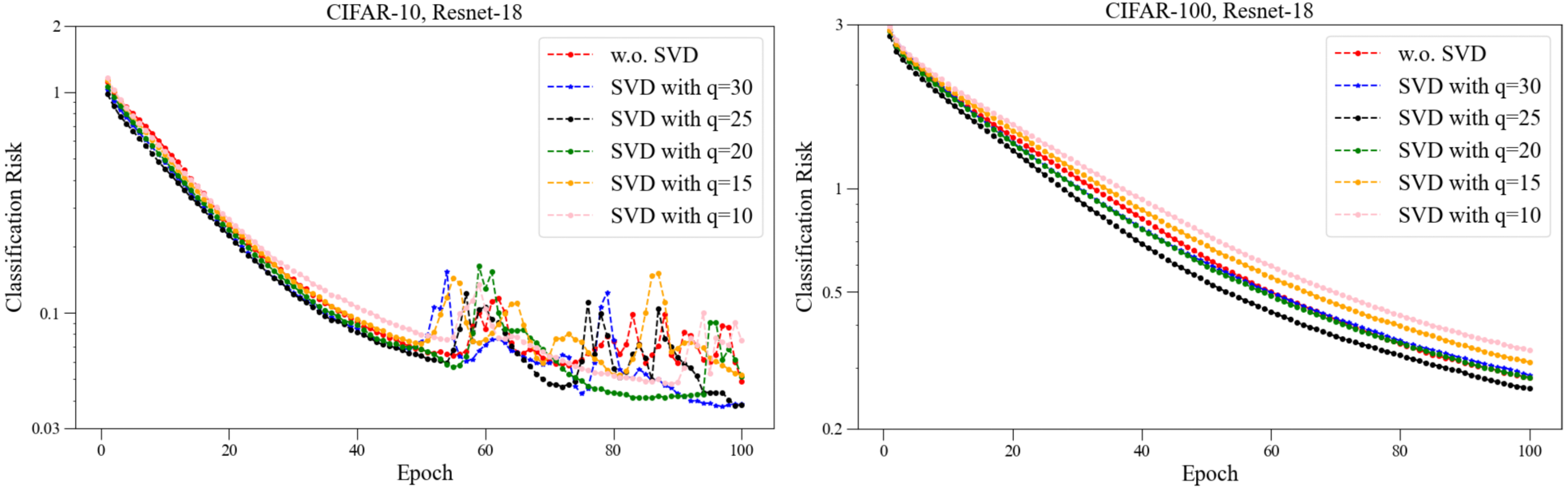}
    \caption{Classification risks of SimCLR using the augmentation $\mathcal{T}_1$ and the truncated SVD with different values of $q$. The corresponding top-1 accuracies are shown in Table \ref{Table of w/o SVD vs. SVD under different q}.}
    \label{Classification risk of CIFAR10Resnet18 + CIFAR100Resnet18}
\end{figure*}

\begin{table}[ht]
\caption{Downstream classification top-1 accuracies ($\%$) of SimCLR ($\mathcal{L}_{InfoNCE}$) using the truncated SVD ($q=30$ for CIFAR-10 and CIFAR-100, $q=90$ for STL-10) with different embedding dimension $k$.}
\label{Table of different k}
    \centering
    \renewcommand\arraystretch{1.5}
    \begin{tabular}{c|c|c|ccccc}
        \hline
        \multirow{2}{*}{$\mathcal{T}$} & \multirow{2}{*}{Encoder} & \multirow{2}{*}{Dataset} & \multicolumn{5}{c}{Embedding Dimension}\\
        \cline{4-8}
        & & & $k=128$ & $k=256$ & $k=512$ & $k=1024$ & $k=2048$\\
        \hline
        $\mathcal{T}_1$ & Resnet-18 & CIFAR-10 & 68.12 & 69.11 & \textbf{69.48} & 69.27 & 68.84\\
        $\mathcal{T}_1$ & Resnet-50 & CIFAR-10 & 67.72 & 68.29 & \textbf{69.12} & 67.83 & 63.36\\
        $\mathcal{T}_1$ & Resnet-152 & CIFAR-10 & 66.75 & 67.64 & \textbf{68.05} & 65.98 & 62.50\\
        $\mathcal{T}_1$ & Resnet-18 & CIFAR-100 & 37.68 & 38.49 & \textbf{39.59} & 39.42 & 39.19\\
        $\mathcal{T}_1$ & Resnet-50 & CIFAR-100 & 38.96 & 39.50 & \textbf{39.83} & 38.10 & 32.23\\
        $\mathcal{T}_1$ & Resnet-18 & STL-10 & 71.28 & 71.37 & \textbf{71.93} & 71.49 & 71.34\\
        $\mathcal{T}_1$ & Resnet-50 & STL-10 & 72.72 & \textbf{73.76} & 72.53 & 71.73 & 70.28\\
        \hline
    \end{tabular}
\end{table}

\begin{table}[ht]
\caption{Downstream classification top-1 accuracies ($\%$) of SimCLR ($\mathcal{L}_{InfoNCE}$) using the truncated SVD ($q=30$ for CIFAR-10) with different epochs.}
\label{Table of different epoch}
    \centering
    \renewcommand\arraystretch{1.5}
    \begin{tabular}{c|c|c|c|ccccc}
        \hline
        \multirow{2}{*}{$\mathcal{T}$} & \multirow{2}{*}{Encoder} & \multirow{2}{*}{Dataset} & \multirow{2}{*}{SVD} & \multicolumn{5}{c}{Epochs}\\
        \cline{5-9}
        & & & & 100 & 200 & 300 & 400 & 500\\
        \hline
        $\mathcal{T}_1$ & Resnet-18 & CIFAR-10 & w.o. SVD & 68.82 & 70.91 & 71.05 & 72.97 & 74.54\\
        $\mathcal{T}_1$ & Resnet-18 & CIFAR-10 & $q=30$ & \textbf{69.48} & \textbf{71.06} & \textbf{71.48} & \textbf{73.43} & \textbf{74.94}\\
        \hline
    \end{tabular}
\end{table}

\begin{table}[!ht]
\caption{Downstream classification top-1 accuracies ($\%$) of MoCo ($\mathcal{L}_{InfoNCE}$) using the truncated SVD with different $q$.}
\label{Table of w/o SVD vs. SVD under different q for MoCo}
    \centering
    \renewcommand\arraystretch{1.5}
    \begin{tabular}{c|c|c|cccccc}
        \hline
        $\mathcal{T}$ & Encoder & Dataset & w/o SVD & $q=30$ & $q=25$ & $q=20$ & $q=15$ & $q=10$\\
        \hline
        $\mathcal{T}_1$ & Resnet-18 & CIFAR-10 & 72.69 & \textbf{73.02} & 72.48 & 71.58 & 71.30 & 69.87\\
        \hline
    \end{tabular}
\end{table}

\begin{table*}[!ht]
    \centering
    \caption{Downstream classification top-1 accuracies ($\%$) of SimCLR ($\mathcal{L}_{InfoNCE}$) using the truncated SVD on TinyImageNet-200 (image size $64\times 64$, $50$ pre-training epochs) and different backbones (ViT, ConvNeXt, 10 pre-training epochs) with different truncated parameter $q$.}
    \renewcommand\arraystretch{1.5}
    \begin{tabular}{c|c|c|cccccc}
        \hline
        $\mathcal{T}$ & Encoder & Dataset & w/o SVD & $q=60$ & $q=50$ & $q=40$\\
        \hline
        $\mathcal{T}_1$ & Resnet-18 & TinyImageNet-200 & 28.72 & \textbf{29.38} & 28.44 & 27.94\\
        $\mathcal{T}_1$ & ViT & CIFAR-10 & 42.22 & \textbf{42.98} & 42.87 & 38.44\\
        $\mathcal{T}_1$ & ConvNeXt & CIFAR-10 & 55.75 & \textbf{56.37} & 55.87 & 55.59\\
        \hline
    \end{tabular}
\end{table*}

\begin{table*}[!ht]
\caption{Downstream classification top-1 accuracies $(\%)$ of SimCLR ($\mathcal{L}_{InfoNCE}$) on CIFAR-10 using the truncated SVD with Random Erasing, GridMask and HidePatch.}
    \centering
    \setlength{\tabcolsep}{1mm}
    \renewcommand\arraystretch{1.5}
    \begin{tabular}{c|c|ccc}
        \hline
        SVD & Encoder & Random Erasing & GridMask & HidePatch\\
        \hline
        w.o. SVD & Resnet-18 & 49.27 & 50.33 & 23.05\\
        $q=30$ & Resnet-18 & \textbf{49.65} & \textbf{51.20} & \textbf{26.22}\\
        \hline
    \end{tabular}
\end{table*}

\begin{table}[ht]
\caption{Downstream classification top-1 accuracies ($\%$) of SimCLR ($\mathcal{L}_{InfoNCE}$, 10 pre-training epochs) on ViT and ConvNeXt using the truncated SVD ($q=30$ for CIFAR-10) with different embedding dimension $k$ ($-$ represents that it does not converge).}
    \centering
    \renewcommand\arraystretch{1.5}
    \begin{tabular}{c|c|c|ccccccc}
        \hline
        \multirow{2}{*}{$\mathcal{T}$} & \multirow{2}{*}{Encoder} & \multirow{2}{*}{Dataset} & \multicolumn{7}{c}{Embedding Dimension}\\
        \cline{4-10}
        & & & $k=128$ & $k=256$ & $k=512$ & $k=1024$ & $k=2048$ & $k=3072$ & $k=4096$\\
        \hline
        $\mathcal{T}_1$ & ViT & CIFAR-10 & 41.62 & 41.28 & 42.22 & \textbf{43.14} & 40.18 & $-$ & $-$\\
        $\mathcal{T}_1$ & ConvNeXt & CIFAR-10 & 53.48 & 55.58 & 55.75 & 55.89 & \textbf{56.71} & 55.84 & 55.13\\
        \hline
    \end{tabular}
\end{table}
\begin{table}[ht]
\caption{Downstream classification top-1 accuracies ($\%$) of BYOL ($\mathcal{L}_{InfoNCE}$, 10 pre-training epochs) using the truncated SVD ($q=30$ for CIFAR-10) with different embedding dimension $k$.}
    \centering
    \renewcommand\arraystretch{1.5}
    \begin{tabular}{c|c|c|ccccccc}
        \hline
        \multirow{2}{*}{$\mathcal{T}$} & \multirow{2}{*}{Encoder} & \multirow{2}{*}{Dataset} & \multicolumn{7}{c}{Embedding Dimension}\\
        \cline{4-10}
        & & & $k=128$ & $k=256$ & $k=512$ & $k=1024$ & $k=2048$ & $k=3072$ & $k=4096$\\
        \hline
        $\mathcal{T}_1$ & BYOL & CIFAR-10 & 33.71 & 33.85 & 34.08 & 34.09 & \textbf{34.32} & 33.11 & 33.00\\
        \hline
    \end{tabular}
\end{table}

\begin{table}[ht]
\caption{Downstream classification top-1 accuracies ($\%$) of SimCLR ($\mathcal{L}_{InfoNCE}$) on CIFAR-100 and STL-10 using the truncated SVD with different $q$ or the data inflation strategy under the weak data augmentation adopted by \citet{DBLP:conf/iclr/WangZ024} $(\mathcal{T}_8=\{$RRC(0.2, 1.0), Color jitter(0.5, 0.4), Random horizontal flip, Random grayscale, Gaussian blur$\})$.}
    \centering
    \renewcommand\arraystretch{1.5}
    \begin{tabular}{c|c|c|ccccc}
        \hline
        Dataset & $\mathcal{T}$ & Encoder & Inflation & w/o SVD & $q=90$ & $q=70$ & $q=50$\\
        \hline
        STL-10 & $\mathcal{T}_8$ & Resnet-18 & 70.86 & 70.51 & \textbf{71.49} & 71.23 & 69.26\\
        \hline
        \hline
        Dataset & $\mathcal{T}$ & Encoder & Inflation & & Inflation + $(q=90)$ & + $(q=70)$ & + $(q=50)$\\
        \hline
        STL-10 & $\mathcal{T}_8$ & Resnet-18 & 70.86 & & \textbf{71.75} & 71.21 & 69.85\\
        \hline
        \hline
        Dataset & $\mathcal{T}$ & Encoder & Inflation & w/o SVD & $q=30$ & $q=25$ & $q=20$\\
        \hline
        CIFAR-100 & $\mathcal{T}_8$ & Resnet-18 & 42.84 & 42.29 & \textbf{42.93} & 42.72 & 42.5\\
        \hline
        \hline
        Dataset & $\mathcal{T}$ & Encoder & Inflation & & Inflation + $(q=30)$ & + $(q=25)$ & + $(q=20)$\\
        \hline
        CIFAR-100 & $\mathcal{T}_8$ & Resnet-18 & 42.84 & & 43.04 & \textbf{43.08} & 42.92\\
        \hline
    \end{tabular}
\end{table}

\begin{table}[ht]
\caption{The estimated frequencies ($\%$) of labeling error for CIFAR-10 with multiply augmentation strategies.}
    \centering
    \renewcommand\arraystretch{1.5}
    \begin{tabular}{c|cccc}
        \hline
        Augmentation & RRC & $\mathcal{T}_1$ & $\mathcal{T}_2$ & $\mathcal{T}_3$\\
        \hline
        Frequency & 27.9 & 37.5 & 48.3 & 42.1\\
        \hline
    \end{tabular}
\end{table}

\begin{table}[ht]
\caption{The cost of SVD on different datasets.}
    \centering
    \renewcommand\arraystretch{1.5}
    \begin{tabular}{c|ccc}
        \hline
        Dataset & CIFAR-10 & TinyImageNet-200 & STL-10\\
        \hline
        Size & $32\times 32$ & $64\times 64$ & $96\times 96$\\
        \hline
        Truncated parameter $q$ & $q=30$ & $q=60$ & $q=90$\\
        \hline
        Time of per image / s & 0.001994 & 0.004985 & 0.007978\\
        Time of all images / s & 99.7186 & 498.5094 & 797.7724\\
        \hline
    \end{tabular}
\end{table}

\end{document}